\documentclass[final]{cvpr}

\usepackage{times}
\usepackage{epsfig}
\usepackage{graphicx}
\usepackage{amsmath}
\usepackage{amssymb}
\usepackage{subfigure}
\usepackage{graphicx}
\usepackage{caption}
\usepackage{longtable}
\usepackage{multirow}
\usepackage{algorithm}
\usepackage{algorithmicx}
\usepackage{algpseudocode}
\usepackage{amsmath}
\usepackage{indentfirst} 
\usepackage{verbatim}
\usepackage{color} 
\usepackage{verbatim}
\usepackage{appendix}
\usepackage{threeparttable}
\usepackage{tablefootnote}
\usepackage{url}
\usepackage{diagbox}
\usepackage[misc]{ifsym} 

\usepackage[pagebackref=true,breaklinks=true,colorlinks,bookmarks=false]{hyperref}

\def\cvprPaperID{1456} 
\def\confYear{CVPR 2021}

\begin{document}

\title{Farewell to Mutual Information: Variational Distillation for Cross-Modal Person Re-Identification}

\author{Xudong Tian$^1$, Zhizhong Zhang$^1$$^{(\textrm{\Letter})}$, Shaohui Lin$^1$, Yanyun Qu$^2$, Yuan Xie$^1$$^{(\textrm{\Letter})}$, Lizhuang Ma$^{1,3}$\\
	$^1$East China Normal University, $^2$Xiamen University, $^3$Shanghai Jiao Tong University\\
	{\tt\small txd51194501066@gmail, zzzhang@cs.ecnu.edu.cn, shaohuilin007@gmail.com,}\\ {\tt\small yyqu@xmu.edu.cn, xieyuan8589@foxmail.com, ma-lz@cs.sjtu.edu.cn}
}

\maketitle


\begin{abstract}
	The Information Bottleneck (IB) provides an information theoretic principle for representation learning, by retaining all information relevant for predicting label while minimizing the redundancy. Though IB principle has been applied to a wide range of applications, its optimization remains a challenging problem which heavily relies on the accurate estimation of mutual information. In this paper, we present a new strategy, Variational Self-Distillation (VSD), which provides a scalable, flexible and analytic solution to essentially fitting the mutual information but without explicitly estimating it. Under rigorously theoretical guarantee, VSD enables the IB to grasp the intrinsic correlation between representation and label for supervised training. Furthermore, by extending VSD to multi-view learning, we introduce two other strategies, Variational Cross-Distillation (VCD) and Variational Mutual-Learning (VML), which significantly improve the robustness of representation to view-changes by eliminating view-specific and task-irrelevant information. To verify our theoretically grounded strategies, we apply our approaches to cross-modal person Re-ID, and conduct extensive experiments, where the superior performance against state-of-the-art methods are demonstrated. Our intriguing findings highlight the need to rethink the way to estimate mutual information. 
	
\end{abstract}

\section{Introduction}

The Information Bottleneck (IB) \cite{ib} has made remarkable progress in the development of modern machine perception systems such as computer vision \cite{mib}, speech processing \cite{speechprocessing}, neuroscience \cite{neuroscience} and natural language processing \cite{nlp}. It is essentially an information-theoretic principle that transforms raw observation into a, typically lower-dimensional, representation and this principle is naturally extended to representation learning
or understanding Deep Neural Networks (DNNs) \cite{understandDNNs_1, understandDNNs_2,understandDNNs_3}.

By fitting mutual information (MI), IB allows the learned representation to preserve complex intrinsic correlation structures over high dimensional data and contain the information relevant with downstream task \cite{ib}. However, despite successful applications, there is a significant drawback in
conventional IB hindering its further development ({\it i.e.,} estimation of mutual information). 

In practice, mutual information is a fundamental quantity for measuring the statistical dependencies \cite{mine} between random variables, but its estimation remains a challenging problem. To address this, traditional approaches \cite{non-para_MIE_1,non-para_MIE_2,non-para_MIE_3,non-para_MIE_4,non-para_MIE_5,non-para_MIE_6,non-para_MIE_7,non-para_MIE_8,non-para_MIE_9,non-para_MIE_10} mainly resort to non-parametric estimator under very limited problem setting where the probability distributions are known or the variables are discrete \cite{variationalbound,mine}. To overcome this constraint, some works \cite{infomax,mib,variationalbound} adopt the trainable parametric neural estimators involving reparameterization, sampling, estimation of posterior distribution \cite{mine}, which, unfortunately, practically has very poor scalability. Besides, the estimation of posterior distribution would become intractable when the network is complicated.

Another obvious drawback of the conventional IB is that, the optimization of IB is essentially a tradeoff between having a concise representation and one with good predictive power, which makes it impossible to achieve both high compression and accurate prediction \cite{ib,vib,dualIB,DMIB}. Consequently, the optimization of conventional IB becomes tricky, and its robustness is also seriously compromised due to the mentioned reasons.

In this paper, we propose a new strategy for the information bottleneck named as Variational Self-Distillation (VSD), which enables us to preserve sufficient task-relevant information while simultaneously discarding task-irrelevant distractors. We should emphasize here that our approach essentially \textbf{fits the mutual information but without explicitly estimating it}. To achieve this, we use variational inference to provide a theoretical analysis which obtains an analytical solution to VSD. Different from traditional methods that attempt to develop estimators for mutual information, our method avoids all complicated designs and allows the network to grasp the intrinsic correlation between the data and label with theoretical guarantee.

Furthermore, by extending VSD to multi-view learning, we propose Variational Cross-Distillation (VCD) and Variational Mutual-Learning (VML), the strategies that improve the robustness of information bottleneck to view-changes. VCD and VML eliminate the view-specific and task-irrelevant information without relying on any strong prior assumptions. More importantly, we implement VSD, VCD and VML in the form of training losses and they can benefit from each other, boosting the performance. As a result, two key characteristics of representation learning ({\it i.e.,} sufficiency and consistency) are kept by our approach.

To verify our theoretically grounded strategies, we apply our approaches to cross-modal person re-identification\footnote{We don't explicitly distinguish multi-view and multi-modal throughout this paper.}, a cross-modality pedestrian image matching task. Extensive experiments conducted on the widely adopted benchmark datasets demonstrate the effectiveness, robustness and impressive performance of our approaches against state-of-the-arts methods. Our main contributions are summarized as follows:
\begin{itemize}
	\item We design a new information bottleneck strategy (VSD) for representation learning. By using variational inference to reconstruct the objective of IB, we can preserve sufficient label information while simultaneously getting rid of task-irrelevant details.
	
	\item A scalable, flexible and analytical solution to fitting mutual information is presented through rigorous theoretical analysis, which fundamentally tackle the difficulty of estimation of mutual information. 
	
	\item We extend our approach to multi-view representation learning, and it significantly improve the robustness to view-changes by eliminating the view-specific and task-irrelevant information.
	
\end{itemize}

\section{Related Work and Preliminaries}\label{preliminaries}

The seminal work is from \cite{ib}, which introduces the IB principle. On this basis, \cite{vib,mib,vdb} either reformulate the training objective or extend the IB principle, remarkably facilitating its application.

In contrast to all of the above, our work is the first to provide an analytical solution to fitting the mutual information without estimating it. The proposed VSD can better preserve task-relevant information, while simultaneously getting rid of task-irrelevant nuisances. Furthermore, we extend VSD to multi-view setting and propose VCD and VML with significantly promoted robustness to view-changes.

To better illustrate, we provide a brief review of the IB principle \cite{ib} in the context of supervised learning. Given data observations $V$ and labels $Y$, the goal of representation learning is to obtain an encoding $Z$ which is maximally informative to $Y$, measured by mutual information:
\begin{equation}
	I(Z;Y)=\int  p(z,y)\log\frac{p(z,y)}{p(z) p(y)}dzdy.
\end{equation} 

To encourage the encoding process to focus on the label information, IB was proposed to enforce an upper bound $I_c$ to the information flow from the observations $V$ to the encoding $Z$, by maximizing the following objective:
\begin{equation}\label{definition of ib}
	\max I(Z;Y) ~ s.t.~I(Z;V)\le I_c.
\end{equation}

Eq. (\ref{definition of ib}) implies that a compressed representation can improve the generalization ability by ignoring irrelevant distractors in the original input.
By using a Lagrangian objective, IB allows the encoding $Z$ to be maximally expressive about $Y$ while being maximally compressive about $X$ by:
\begin{equation}\label{objective of ib}
	\mathcal{L}_{IB}=I(Z;V)-\beta I(Z;Y),
\end{equation}
where $\beta$ is the Lagrange multiplier. However, it has been shown that it is impossible to achieve both objectives in Eq. (\ref{objective of ib}) practically \cite{mib,vib} due to the trade-off optimization between high compression and high mutual information. 

More significantly, estimating mutual information in high dimension imposes additional difficulties \cite{mie_1,mine,variationalbound} for optimizing IB. As a consequence, it inevitably introduces irrelevant distractors and discards some predictive cues in the encoding process. Next, we show how we design a new strategy to deal with these issues, and extend it to multi-view representation learning.


\section{Method}
Let $v\in V$ be an observation of input data $x \in X$ extracted from an encoder $E(v|x)$. The challenge of optimizing an information bottleneck can be formulated as finding an extra encoding $E(z|v)$ that preserves all label information contained in $v$, while simultaneously discarding task-irrelevant distractors. To this end, we show the key roles of two characteristics of $z$, ({\it i.e.}, \textbf{sufficiency} and \textbf{consistency}) based on the information theory, and design two variational information bottlenecks to keep both characteristics.


To be specific, we propose a Variational Self-Distillation (VSD) approach, which allows the information bottleneck to preserve sufficiency of representation $z$, where the amount of label information is unchanged after the encoding process. In the design of VSD, we further find it can be extended to multi-view task, and propose Variational Cross-Distillation (VCD) and Variational Mutual-Learning (VML) approaches based on the consistency of representations, both of which are able to eliminate the sensitivity of view-changes and improve the generalization ability for multi-view learning. More importantly, the proposed VSD, VCD and VML can benefit from each other, and essentially fit the mutual information in the high dimension without explicitly estimating it through the theoretical analysis.

\subsection{Variational Self-Distillation}\label{vsd}
An information bottleneck is used to produce representations $z$ for keeping all predictive information {\it w.r.t} label $y$ while avoiding encoding task-irrelevant information. It is also known as sufficiency of $z$ for $y$, which is defined as:
\begin{equation}\label{definition of sufficiency}
	I(z;y)=I(v;y),
\end{equation}
where $v$ is an observation containing all label information.
By factorizing the mutual information between $v$ and $z$, we can identify two terms \cite{mib}:
\begin{equation}
	I(v;z)=I(z;y)+I(v;z|y),\label{factorization1}
\end{equation}
where $I(z;y)$ denotes the amount of label information maintained in the representation $z$, and $I(v;z|y)$ represents the irrelevant information encoded in $z$ regarding given task \cite{mib}, {\it i.e.}, superfluous information. Thus sufficiency of $z$ for $y$ is formulated as maximizing $I(z;y)$ and simultaneously minimizing $I(v;z|y)$. Based on the data processing inequality $I(z;y)\le I(v;y)$, we have:
\begin{equation}
	I(v;z)\le I(v;y)+I(v;z|y).\label{ineq1}
\end{equation}
The first term in the right of Eq. (\ref{ineq1}) indicates that preserving sufficiency undergoes two sub-processes: maximizing $I(v;y)$, and forcing $I(z;y)$ to approximate $I(v;y)$.

In this view, sufficiency of $z$ for $y$ is re-formulated as three sub-optimizations: maximizing $I(v;y)$, minimizing $I(v;y)-I(z;y)$ and minimizing $I(v;z|y)$. Obviously, maximizing the first term $I(v;y)$ is strictly consistent with the specific task and the last two terms are equivalent. Hence the optimization is simplified to:
\begin{equation}\label{obj1}
	\min I(v;y)-I(z;y).
\end{equation}
However, it is hard to conduct min-max game in Eq. (\ref{factorization1}), due to the notorious difficulty in estimating mutual information in high dimension, especially when involving latent variable optimization. To deal with this issue, we introduce the following theory:

~\\
\noindent{\textbf{Theorem 1. }}{\textit{Minimizing Eq. (\ref{obj1}) is equivalent to minimizing the subtraction of conditional entropy $H(y|z)$ and $H(y|v)$. That is:
		\begin{flalign}
			&\min I(v;y)-I(z;y)\iff \min H(y|z)-H(y|v),\nonumber
		\end{flalign} 
		where $H(y|z):=-\int p(z)dz\int p(y|z)\log p(y|z)dy$.}} 
~\\

More specifically, given a sufficient observation $v$ for $y$, we have following Corollary:

~\\
\noindent{\textbf{Corollary 1.} \textit{If the KL-divergence between the predicted distributions of a sufficient observation $v$ and the representation $z$ equals to $0$, then $z$ is sufficient for $y$ as well {\it i.e.},
		\begin{equation}
			D_{KL}[\mathbb{P}_v||\mathbb{P}_z]=0 \implies H(y|z)-H(y|v)=0, \nonumber
		\end{equation}
		where $\mathbb P_z=p(y|z)$, $\mathbb P_v=p(y|v)$ represent the predicted distributions, and $D_{KL}$ denotes the KL-divergence.}}
~\\

Detailed proof and formal assertions can be found in supplementary material.
As a consequence, sufficiency of $z$ for $y$ could be achieved by the following objective:
\begin{equation}
	\mathcal{L}_{VSD}=\min _{\theta, \phi} \mathbb{E}_{v \sim E_{\theta}(v|x)}\left[\mathbb{E}_{z \sim E_{\phi}(z|v)} \left[D_{K L}[\mathbb P_v \|\mathbb  P_z]\right]\right],\label{obj of vsd}
\end{equation}

\begin{figure}[t]
	\centering
	\includegraphics[width=0.85\linewidth]{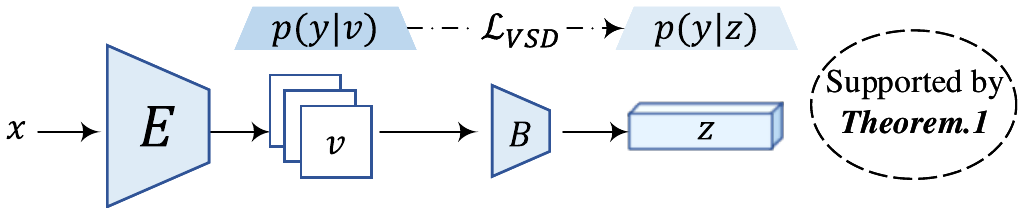}
	\caption{Illustration of VSD, in which $E$ and $B$ denote the encoder and the information bottleneck, respectively.}
	\label{modal-specific branches}
	\vspace{-5mm}
\end{figure}

where $\theta, \phi$ stand for the parameters of the encoder and information bottleneck, respectively. On the other hand, based on Eq. (\ref{ineq1}) and Eq. (\ref{factorization1}), the minimization of $I(v;y)-I(z;y)$ is equivalent to reducing $I(v;z|y)$, indicating that Eq. (\ref{obj of vsd}) also enables IB to eliminate irrelevant distractors. In this perspective, our approach is essentially a self-distillation method that purifies the task-relevant knowledge. 
More importantly, through using the variational inference, we reformulate the objective of IB, and provide a theoretical analysis which obtains an analytical solution to fitting mutual information in high dimension. Hence we name our strategy as Variational Self-Distillation, {\it i.e.}, VSD.


\textbf{Discussion.} Compared with other self-distillation methods like \cite{self-distillation}, one primary advantage of our approach is that VSD is able to retrieve those useful but probably discarded information while simultaneously avoiding task-irrelevant information under theoretically guaranteed. Different from explicitly reducing $I(v;z)$, we iteratively perform VSD to make the representation sufficient for the task. Ideally, when we have $I(v;y)=I(z;y)$, we could achieve the sufficient representation with minimized superfluous information, {\it i.e.}, optimal representation.

\subsection{Variational Cross-Distillation and Variational Mutual-Learning}\label{vcd and vml}
As more and more real-world data are collected from diverse sources or obtained from different feature extractors, multi-view representation learning has gained increasing attention. In this section, we show VSD could be flexibly extended to multi-view learning. 

Consider $v_1$ and $v_2$ as two observations of $x$ from different viewpoints. Assuming that both $v_1$ and $v_2$ are sufficient for label $y$, thus any representation $z$ containing all the information accessible from both views would also contain the necessary label information. More importantly, if $z$ only captures the cues that are accessible from both $v_1$ and $v_2$, it would eliminate the view-specific details and is robust to view-changes \cite{mib}.

Motivated by this, we define {\textbf{consistency}} {\it w.r.t} $z_1,z_2$ obtained from an information bottleneck as:

~\\{\textbf{Definition 1. Consistency:}} \textit{$z_1$ and $z_2$ are view-consistent if and only if $I(z_1;y)=I(v_1v_2;y)=I(z_2;y).$}~\\

Intuitively, $z_1$ and $z_2$ are view-consistent only when they possess equivalent amount of predictive information. Analogous to Eq. (\ref{factorization1}), we first factorize the mutual information between the observation $v_1$ and representation $z_1$ to clearly reveal the essence of consistency:
\begin{equation}
	I(v_1;z_1)=\underbrace{I(v_1;z_1|v_2)}_{\operatorname{view-specific}}+\underbrace{I(z_1;v_2)}_{\operatorname{consistent}}.\label{factorization2}
\end{equation}
Suggested by \cite{mib}, $I(v_1;z_1|v_2)$ represents that, the information contained in $z_1$ is unique to $v_1$ and is not predictable by observing $v_2$, {\it i.e.}, view-specific information, and $I(z_1;v_2)$ denotes the information shared by $z_1$ and $v_2$, which is named as view-consistent information.

To obtain view-consistent representation with minimal view-specific details, we need to jointly minimize $I(v_1;z_1|v_2)$ and maximize $I(z_1;v_2)$. On the one hand, to reduce view-specific information and note that $y$ is constant, we can use the following equation to approximate the upper bound of $I(v_1;z_1|v_2)$ (proofs could be found in supplementary material).
\begin{equation}
	\min _{\theta, \phi}\mathbb{E}_{v_1,v_2 \sim E_{\theta}(v|x)}\mathbb{E}_{z_1,z_2 \sim E_{\phi}(z|v)} \left[D_{K L}[\mathbb P_{z_1} \|\mathbb  P_{z_2}]\right],\label{vml1}
\end{equation}
where $\mathbb P_{z_1}=p(y|z_1)$ and $\mathbb P_{z_2}=p(y|z_2)$ denote the predicted distributions. Similarly, we have the following objective to reduce $I(v_2;z_2|v_1)$:
\begin{equation}
	\min _{\theta, \phi}\mathbb{E}_{v_1,v_2 \sim E_{\theta}(v|x)}\mathbb{E}_{z_1,z_2 \sim E_{\phi}(z|v)} \left[D_{K L}[\mathbb P_{z_2} \|\mathbb  P_{z_1}]\right].\label{vml2}
\end{equation}
By combining Eq. (\ref{vml1}) and Eq. (\ref{vml2}), we introduce the following objective to minimize the view-specific information for both $z_1$ and $z_2$:
\begin{equation}
	\mathcal{L}_{VML}=\min _{\theta, \phi}\mathbb{E}_{v_1,v_2 \sim E_{\theta}(v|x)}\mathbb{E}_{z_1,z_2 \sim E_{\phi}(z|v)} \left[D_{JS}[\mathbb P_{z_1} \|\mathbb  P_{z_2}]\right],\label{vml3}
\end{equation}
in which $D_{JS}$ represents Jensen-Shannon divergence. In practice, Eq. (\ref{vml3}) forces $z_1,z_2$ to learn from each other, thus we name it variational mutual-learning.

On the other hand, by using the chain rule to subdivide $I(z_1;v_2)$ into two components \cite{mib}, we have:

\begin{equation}
	I(z_1;v_2)=\underbrace{I(v_2;z_1|y)}_{\operatorname{superfluous}}+\underbrace{I(z_1;y)}_{\operatorname{predictive}}.\label{factorization3}
\end{equation}
Combining with Eq. (\ref{factorization2}), we have:
\begin{equation}\label{factorization4}
	I(v_1;z_1)=\underbrace{I(v_1;z_1|v_2)}_{\operatorname{view-specific}}+\underbrace{I(v_2;z_1|y)}_{\operatorname{superfluous}}+\underbrace{I(z_1;y)}_{\operatorname{predictive}}.
\end{equation}
Eq. (\ref{factorization4}) implies that the view-consistent information also includes superfluous information. Therefore, based on the above analysis, we give the following theorem to clarify view-consistency:

\begin{figure}[t]
	\centering
	\includegraphics[width=0.85\linewidth]{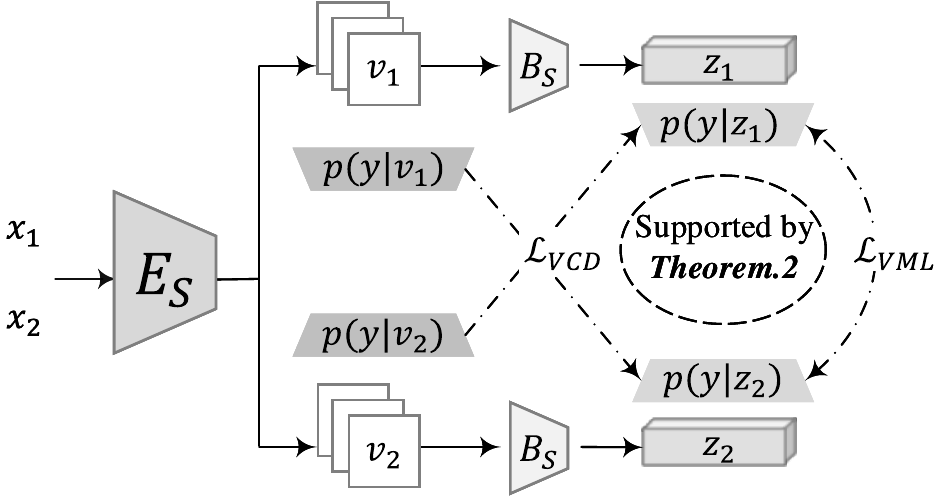}
	\caption{Illustration of $\mathcal{L}_{VCD}$ and $\mathcal{L}_{VML}$, where the subscripts are used to denote different views and $E_S,B_S$ refer to the parameter-shared encoder and information bottleneck.}
	\label{modal-shared branch}
	\vspace{-6mm}
\end{figure}

~\\
\noindent{\textbf{Theorem 2.}} {\textit{Given two different sufficient observations $v_1,v_2$ of an input $x$, the corresponding representations $z_1$ and $z_2$ are view-consistent when the following conditions are satisfied:}:
	\begin{flalign}\nonumber
		& I(v_1;z_1|v_2)+I(v_2;z_1|y) \le 0,\\ \nonumber
		& I(v_2;z_2|v_1)+I(v_1;z_2|y) \le 0. \nonumber
	\end{flalign} 
}
The proof of Theorem 2 can be found in supplementary material.

According to Theorem 1 and Corollary 1, the following objective can be introduced to promote consistency between $z_1$ and $z_2$:
\begin{equation}
	\mathcal{L}_{VCD}=\min _{\theta, \phi} \mathbb{E}_{v_1,v_2 \sim E_{\theta}(v|x)}\mathbb{E}_{z_1,z_2 \sim E_{\phi}(z|v)} \left[D_{K L}[\mathbb P_{v_2} \|\mathbb  P_{z_1}]\right],\label{vcd}
\end{equation}
where $\mathbb P_{z_1}=p(y|z_1)$ and $\mathbb P_{v_2}=p(y|v_2)$ denote the predicted distributions. Based on Theorem 1 and Corollary 1, Eq. (\ref{vcd}) enables the representation $z_1$ to preserve predictive cues, while simultaneously eliminating the superfluous information contained in $I(z_1;v_2)$ (vice versa for $z_2$ and $I(z_2;v_1)$), and is named as variational cross-distillation.

\textbf{Discussion.} Notice that MIB \cite{mib} is also a multi-view information bottleneck approach. However, there are three fundamental differences between ours and MIB: 1) our strategy essentially fits the mutual information without estimating it through the variational inference. 2) our method does not rely on the strong assumption proposed in \cite{mib} that each view provides the same task-relevant information. Instead, we explore both the complementarity and consistency of multiple views for representation learning. 3) MIB is essentially an unsupervised method and hence it keeps all consistent information in various views due to the lack of label supervision. However, with the predictive information, our method is able to discard superfluous information contained in consistent representation, and hence show improved robustness.

\subsection{Multi-Modal Person Re-ID}
In this section, we show how we apply VSD, VCD and VML to multi-modal learning ({\it i.e.}, Multi-Modal Person Re-ID). In this context, there are two kinds of images from different modals ({\it i.e.}, infrared images $x_I$ and visible images $x_V$). The essential objective of Multi-Modal Person Re-ID is to match the target person from a gallery of images from another modal. 

In particular, we use two parallel modal-specific branches equipped with VSD to handle the images from a specific modal. Besides, as shown in Fig. \ref{reid framework}, a modal-shared branch trained with VCD and VML is deployed to produce modal-consistent representations. To facilitate Re-ID learning, we also adopt some commonly-used strategies in the Re-ID community. Thus the overall loss is given as:
\begin{equation}\label{overall loss}
	\mathcal{L}_{train} = \mathcal{L}_{ReID} + \beta\cdot (\mathcal{L}_{VSD}+\mathcal{L}_{VCD} + \mathcal{L}_{VML}).
\end{equation}
More specifically, $\mathcal{L}_{ReID}$ can be further divided into the following terms,
\begin{equation}\label{reid loss}
	\mathcal{L}_{ReID}=\mathcal{L}_{cls}+\mathcal{L}_{metric}+\alpha\cdot\mathcal{L}_{DML},
\end{equation}
where $\mathcal{L}_{cls}$, $\mathcal{L}_{metric}$, $\mathcal{L}_{DML}$ denote the classification loss with label smooth \cite{smooth}, metric constraint \cite{rank_loss} and deep mutual learning loss \cite{DML}.




\section{Experiments}\label{experiments}
In this section, we conduct a series of experiments to present a comprehensive evaluation of the proposed methods. To promote the culture of reproducible research, source codes (implemented in MindSpore) will be released later.

\subsection{Experimental settings}
\textbf{Dataset}: The proposed method is evaluated on two benchmarks, {\it i.e.}, SYSU-MM01 \cite{zero-padding} and RegDB \cite{RegDB}. Specifically, SYSU-MM01 is a widely adopted benchmark dataset for infrared-visible person re-identification. It is collected from 6 cameras of both indoor and outdoor environments. It contains 287,628 visible images and 15,792 infrared images of 491 different persons in total, each of which is at least captured by two cameras. RegDB is collected from two aligned cameras (one visible and one infrared) and it totally includes 412 identities, where each identity has 10 infrared images and 10 visible images.	

\begin{figure}[t]
	\centering
	\vspace{-3mm}
	\includegraphics[width=0.85\linewidth]{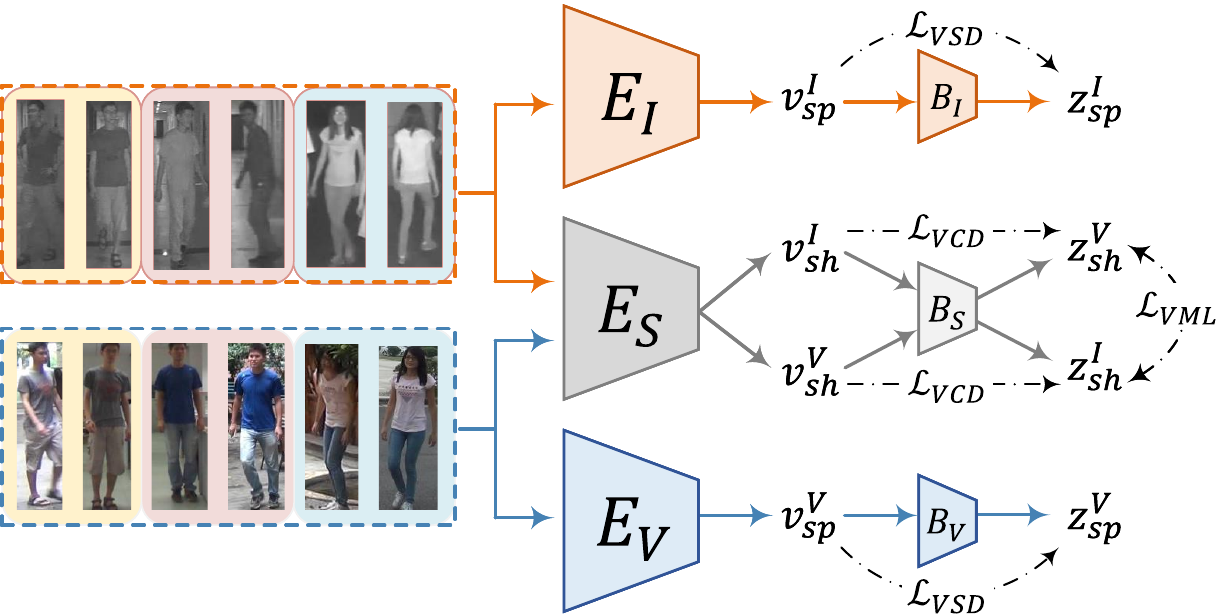}
	\caption{Network architecture for Multi-Modal Re-ID. $E_{I/S/V}$ and $B_{I/S/V}$ represent the encoder (ResNet-50) and information bottleneck (multi-layer perceptrons), respectively. $v$ and $z$ denote the observations and representations from encoder and information bottleneck, respectively}\label{reid framework}
	\vspace{-5mm}
\end{figure}

\textbf{Evaluation Metric}: On both datasets, we follow the most popular protocol \cite{ddag,cm-ssft,xmodality,JSIA-reid} for evaluation, where cumulative match characteristic (CMC) and mean average precision (mAP) are used. On SYSU-MM01, there are two search modes, {\it i.e.}, all-search mode and indoor-search mode. For all-search mode, the gallery consists of all RGB images (captured by CAM1, CAM2, CAM4, CAM5) and the probe consists of all infrared images (captured by CAM3, CAM6). For indoor-search mode, the difference is the gallery only contains images captured from indoor scene (excluding CAM4 and CAM5). On both search modes, we evaluate our model under the single-shot setting \cite{ddag,cm-ssft}. 
On RegDB, following \cite{BDTR}, we report the average result by randomly splitting of training and testing set 10 times. 

\begin{table*}[hbt]
	\centering
	\renewcommand{\arraystretch}{1.0}
	\scriptsize
	\caption{Performance of the proposed method compared with state-of-the-arts. Note that all methods are measured by CMC and mAP on SYSU-MM01 under single-shot mode.}\label{comparision1}
	\vspace{-1.6mm}
	\begin{tabular}{l|c|c|c|c|c|c|c|c|c|c}
		\hline
		\multicolumn{3}{c|}{Settings} & \multicolumn{4}{c|}{All Search}                                   & \multicolumn{4}{c}{Indoor Search}                                \\ \hline
		\multicolumn{1}{c|}{Type}    &	Method    & Venue      & Rank-1 & Rank-10 & Rank-20 & mAP   & Rank-1 & Rank-10 & Rank-20 & mAP   \\ \hline
		Network Design  & Zero-Pad \cite{zero-padding}  & ICCV{\color{blue}'17}   & 14.80  & 54.12   & 71.33   & 15.95 & 20.58  & 68.38   & 85.79   & 26.92 \\ 
		Metric Design  & TONE \cite{TONE_HCML}      & AAAI{\color{blue}'18}   & 12.52  & 50.72   & 68.60   & 14.42 & 20.82  & 68.86   & 84.46   & 26.38 \\ 
		Metric Design  & HCML \cite{TONE_HCML}     & AAAI{\color{blue}'18}   & 14.32  & 53.16   & 69.17   & 16.16 & 24.52  & 73.25   & 86.73   & 30.08 \\ 
		Metric Design  & BDTR \cite{BDTR}     & IJCAI{\color{blue}'18}  & 17.01  & 55.43   & 71.96   & 19.66 & -      & -       & -       & -     \\ 
		Network Design  & cmGAN \cite{CMGAN}    & IJCAI{\color{blue}'18}  & 26.97  & 67.51   & 80.56   & 31.49 & 31.63  & 77.23   & 89.18   & 42.19 \\ 
		Metric Design  & D-HSME \cite{D-HSME}   & AAAI{\color{blue}'18}   & 20.68  & 32.74   & 77.95   & 23.12 & -      & -       & -       & -     \\ 
		Generative  & D$^2$LR \cite{D2LR}  & CVPR{\color{blue}'19}   & 28.9   & 70.6    & 82.4    & 29.2  & -      & -       & -       & -     \\ 
		Metric Design  & MAC \cite{MAC}      & MM'{\color{blue}'19} & 33.26  & 79.04   & 90.09   & 36.22 & 36.43  & 62.36   & 71.63   & 37.03 \\ 
		Generative  & AlignGAN \cite{alignGAN} & ICCV{\color{blue}'19}   & 42.4   & 85.0    & 93.7    & 40.7  & 45.9   & 87.6    & 94.4    & 54.3  \\ 
		Generative  & X-modal \cite{xmodality}  & AAAI{\color{blue}'20}   & 49.92  & 89.79   & 95.96   & 50.73 & -      & -       & -       & -     \\ 
		Generative  & JSIA-ReID \cite{JSIA-reid} & AAAI{\color{blue}'20}   & 38.1   & 80.7    & 89.9    & 36.9  & 52.9   & 43.8    & 86.2    & 94.2  \\ 
		Network Design  & cm-SSFT \cite{cm-ssft}  & CVPR{\color{blue}'20}   & 52.4   & -       & -       & 52.1  & -      & -       & -       & -     \\ 
		Network Design  & Hi-CMD \cite{hi-cmd}  & CVPR{\color{blue}'20}   & 34.94   & 77.58       & -       & 35.94  & -      & -       & -       & -     \\ 
		Network Design  & DDAG  \cite{ddag}  & ECCV{\color{blue}'20}   & {\textbf{{\color{blue}54.75}}}  & {\textbf{{\color{blue}90.39}}}   & {\textbf{{\color{blue}95.81}}}   & {\textbf{{\color{blue}53.02}}} & {\textbf{{\color{blue}61.02}}}  & {\textbf{{\color{blue}94.06}}}   & {\textbf{{\color{blue}98.41}}}   & {\textbf{{\color{blue}67.98}}} \\ \hline
		Representation   & ours            & -            & {\textbf{{\color{red}60.02}}} & {\textbf{\textbf{\color{red}{94.18}}}} & {\textbf{{\color{red}{98.14}}}} & {\textbf{{\color{red}{58.80}}}} & {\textbf{{\color{red}{66.05}}}} & {\textbf{{\color{red}{96.59}}}} & {\textbf{{\color{red}{99.38}}}} & {\textbf{{\color{red}{72.98}}}} \\ \hline
	\end{tabular}
	\vspace{-1.6mm}
	\vspace{-0.9mm}
\end{table*}

\subsection{Implementation Details}
{\textbf{Critical Architectures.}} As visualized in Fig. \ref{reid framework}, we deploy three parallel branches, each of which is composed of a ResNet50 backbone \cite{resnet50} and an information bottleneck. Notice that we drop the last fully-connected layer in the backbone and modify the stride as 1 in last block. Following recent multi-modal Re-ID works \cite{cm-ssft,hi-cmd,xmodality,ddag}, we adopt the strong baseline \cite{bot} with some training tricks, {\it i.e.}, warm up \cite{warmup} (linear scheme for first 10 epochs) and label smooth \cite{smooth}. The information bottleneck is implemented with multi-layer perceptrons of 2 hidden ReLU units of size 1,024 and 512 respectively with an output of size 2$\times$256 that parametrizes mean and variance for the two Gaussian posteriors for the conventional IB principle. 

\textbf{Training.} All experiments are optimized by Adam optimizer with an initial learning rate of 2.6$\times$10$^{-4}$. The learning rate decays 10 times at 200 epochs and we totally train 300 epochs. Horizontal flip and normalization are utilized to augment the training images, where the images are resized to $256\times 128$. The batch size is set to 64 for all experiments, in which it contains 16 different identities, and each identity includes 2 RGB images and 2 IR images. For the hyper-parameters, $\alpha$ and $\beta$ in Eq. (\ref{reid loss}) and Eq. (\ref{overall loss}) are fixed to 8, 2 for all experiments. In order to facilitate the comparison between the conventional IB and the proposed methods, Jensen-Shannon $I_{JS}$ \cite{infomax,variationalbound,mib,mine} is introduced in the estimation of mutual information, which has shown better performance.

\begin{table}[t]
	\centering
	\vspace{-1.6mm}
	\renewcommand{\arraystretch}{1.0}
	\scriptsize
	\caption{Comparison with the state-of-the-arts on RegDB dataset under visible-thermal and thermal-visible settings.}\label{comparision2}
	\vspace{-1.6mm}
	\begin{tabular}{l|c|c|c|c|c}
		\hline
		\multicolumn{2}{c|}{Settings} & \multicolumn{2}{c|}{Visible to Thermal}                         & \multicolumn{2}{c}{Thermal to Visible}                         \\ \hline
		Method   &Venue & Rank-1 & mAP  & Rank-1 & mAP  \\ \hline
		Zero-Pad \cite{zero-padding} &ICCV{\color{blue}'17} & 17.8   & 18.9 & 16.6   & 17.8 \\
		HCML  \cite{TONE_HCML}    &AAAI{\color{blue}'18} & 24.4   & 20.1 & 21.7   & 22.2 \\
		BDTR  \cite{BDTR}    &IJCAI{\color{blue}'18} & 33.6   & 32.8 & 32.9   & 32.0 \\
		D-HSME  \cite{D-HSME}  &AAAI{\color{blue}'18} & 50.8   & 47.0 & 50.2   & 46.2 \\
		D$^2$LR \cite{D2LR}  &CVPR{\color{blue}'19} & 43.4   & 44.1 & -      & -    \\
		MAC   \cite{MAC}    &MM'{\color{blue}'19} & 36.4   & 37.0 & 36.2   & 36.6 \\
		AlignGAN \cite{alignGAN} &ICCV{\color{blue}'19} & 57.9   & 53.6 & 56.3   & 53.4 \\
		X-modal \cite{xmodality}  &AAAI{\color{blue}'20} & 62.2   & 60.2 & -      & -    \\
		JSIA-ReID \cite{JSIA-reid} &AAAI{\color{blue}'20} & 48.5   & 49.3 & 48.1   & 48.9 \\
		cm-SSFT  \cite{cm-ssft} &CVPR{\color{blue}'20} & 62.2   & 63.0 & -      & -    \\
		Hi-CMD  \cite{hi-cmd}  &CVPR{\color{blue}'20} & {\textbf{{\color{blue}70.9}}}   & {\textbf{{\color{blue}66.0}}} & -      & -    \\
		DDAG  \cite{ddag}    &ECCV{\color{blue}'20} & 69.3   & 63.5 & {\textbf{{\color{blue}68.1}}}   & {\textbf{{\color{blue}61.8}}} \\ \hline
		ours      &-& {\textbf{{\color{red}{73.2}}}} & {\textbf{{\color{red}{71.6}}}} & {\textbf{{\color{red}{71.8}}}} & {\textbf{{\color{red}{70.1}}}} \\ 
		\hline
	\end{tabular}
	\vspace{-1.6mm}
	\vspace{-1.6mm}
	\vspace{-1.6mm}
\end{table}

\subsection{Comparison with State-of-the-art Methods}
In this section, we demonstrate the effectiveness of our approach against state-of-the-art methods. 
The comparison mainly consists of three categories, {\it i.e.}, multi-branch network, metric and adversary learning based methods. 

As shown in Tab. \ref{comparision1} and Tab. \ref{comparision2}, we are able to draw the following conclusions: 
1) With recent achievements of generative adversarial network (GAN), generative methods \cite{alignGAN,JSIA-reid,xmodality,D2LR}, have dominated this community, compared with metric methods \cite{BDTR,D-HSME,MAC}. However, despite the promising performance, their success partly attributes to sophisticated designs, ({\it e.g.}, image/modality generation, auxiliary branches, local feature learning, attention block) and therefore result in better performance.
2) Our approach significantly outperforms all the competitors on both datasets. More importantly, we do not require any modality generation process \cite{D2LR,alignGAN,JSIA-reid,xmodality}, which implicitly enlarges the training set. Our optimization is quite efficient since the estimation of mutual information are avoided. The complexity comparison would be illustrated in ablation study. 3) Our method is essentially a representation learning method, and is most relevant to the metric methods. However, our method is based on the information theory to extract predictive information for representation learning, rather than the distance between samples. As a result, our method complements with the metric learning and is able to further boost their performance. To our knowledge, this is the first work to explore the informative representation.

\subsection{Ablation Study}\label{ablation}

In this subsection, we conduct ablation study to show the effectiveness of each component in our approach. For a fair comparison, we follow the same architecture to analyze the influence of sufficiency and consistency for cross-modal person Re-ID. 

We first clarify various settings in this study. As shown in Tab. \ref{effectiveness}, ``\textbf{$E_{S/I/V}$}'' denotes that whether
we use model-shared/infrared/visible branch, where we use $\mathcal{L}_{ReID}$ to train each branch. ``\textbf{$B_{S/I/V}$}'' denotes that in each branch whether we utilize the information bottleneck architecture. ``CIBS'' denotes we adopt conventional IB for training. ``VSD'', ``VML'' and ``VCD'' denote our approaches. It is noteworthy that the results reported in this table are achieved by using observation $v$ or representation $z$ alone.

\begin{table*}[hbt]
	\centering
	\renewcommand{\arraystretch}{1.0}
	\scriptsize
	\caption{Accuracy of the representation $z$ and observation $v$ when using different training strategies. Note that we evaluate the model on SYSU-MM01 under all-search single-shot mode.}\label{effectiveness}
	\vspace{-1.6mm}
	\begin{threeparttable}
		\begin{tabular}{c|ccccccc|cc|cc}
			\hline
			\multicolumn{8}{c|}{Settings} &
			\multicolumn{2}{c|}{Accuracy of $v$} &
			\multicolumn{2}{c}{Accuracy of $z$} \\ \hline
			\diagbox  &
			$E_S$ & $E_I$ and $E_V$ & $B_S$ & $B_I$ and $B_V$ & CIBS & VSD & VCD+VML & \multicolumn{1}{c}{Rank-1} & mAP & \multicolumn{1}{c}{Rank-1} & mAP \\ \hline
			1  & $\surd$ & -       & -       & -       & -       & -       & -       & 44.95 & 46.27 & -     & -     \\
			2  & -       & $\surd$ & -       & -       & -       & -       & -       & 47.25 & 48.31 & -     & -     \\
			3  & $\surd$ & $\surd$ & -       & -       & -       & -       & -       & 48.82 & 49.95 & -     & -     \\ \hline
			4  & $\surd$ & -       & $\surd$ & -       & -       & -       & -       & 47.20 & 47.03 & 38.33 & 41.81 \\
			5  & -       & $\surd$ & -       & $\surd$ & -       & -       & -       & 53.27 & 51.92 & 41.15 & 43.60 \\
			6  & $\surd$ & $\surd$ & $\surd$ & $\surd$ & -       & -       & -       & 54.85 & 53.97 & 42.04 & 44.30 \\ \hline
			7 {\color{red}\tnote{$\dagger$}}  & $\surd$ & -       & $\surd$ & -       & $\surd$ & -       & -       & 8.55  & 11.01 & 24.34 & 28.01 \\
			8 {\color{red}\tnote{$\dagger$}} & -       & $\surd$ & -       & $\surd$ & $\surd$ & -       & -       & 7.76  & 9.77  & 28.69 & 32.42 \\
			9 {\color{red}\tnote{$\dagger$}} & $\surd$ & $\surd$ & $\surd$ & $\surd$ & $\surd$ & -       & -       & 8.79  & 11.23 & 30.43 & 33.67 \\ \hline
			10 & $\surd$ & -       & $\surd$ & -       & -       &         & $\surd$ & 49.22 & 48.74 & 41.48 & 43.02 \\
			11 & -       & $\surd$ & -       & $\surd$ & -       & $\surd$ & -       & {\textbf{\color{blue}59.62}} & {\textbf{\color{blue}57.99}} & {\textbf{\color{blue}50.11}} & {\textbf{\color{blue}51.24}} \\
			12 & $\surd$ & $\surd$ & $\surd$ & $\surd$ & - & $\surd$ & $\surd$ &{\textbf{\color{red}60.02}}  & {\textbf{\color{red}58.80}} & {\textbf{\color{red}50.62}} & {\textbf{\color{red}51.55}} \\ \hline
		\end{tabular}
		\begin{tablenotes}
			\footnotesize
			\item[{\color{red}$\dagger$}] Some results are compared for completeness, as conventional IB does not explicitly enforce any constraints to the observation. 
		\end{tablenotes}
	\end{threeparttable}
	\vspace{-1.6mm}
	\vspace{-1.6mm}
	\vspace{-0.9mm}
	
\end{table*}

\textbf{Main Results.} As illustrated in Tab. \ref{effectiveness}, we have the following observations: 

1) Add model-specific branch and concatenate the features could bring obvious performance improvement (see 1$^{\text {st}}$-3$^{\text {rd}}$ row in Tab. \ref{effectiveness}), which achieves 48.82@Rank-1 and 49.95@mAP. This three-branch network can be considered as a new strong baseline. 

2) By using information bottleneck, the performance is further boosted (see 4$^{\text {th}}$-6$^{\text {th}}$ row in Tab. \ref{effectiveness}). With such a simple modification, we arrive at 54.85@Rank-1 and 53.97@mAP, which beats all state-of-the-art methods.

3) It seems that conventional IB has no advantage in promoting the learning of label information (see the 7$^{\text {th}}$-9$^{\text {th}}$ row in Tab. \ref{effectiveness}). We conjecture it is because that by explicitly reducing $I(v;z)$, conventional IB may not distinguish predictive and superfluous information and probably discard all of them. On the other hand, estimation of mutual information in high dimension appears to be difficult, leading to inaccurate results, especially when involving multi-modal data and latent variables in our setting.

\begin{figure}[t]
	\centering
	\vspace{-1.6mm}
	\subfigure[Accuracy of $z$ when the dimension of IB varies.]{
		\includegraphics[width=0.365\linewidth]{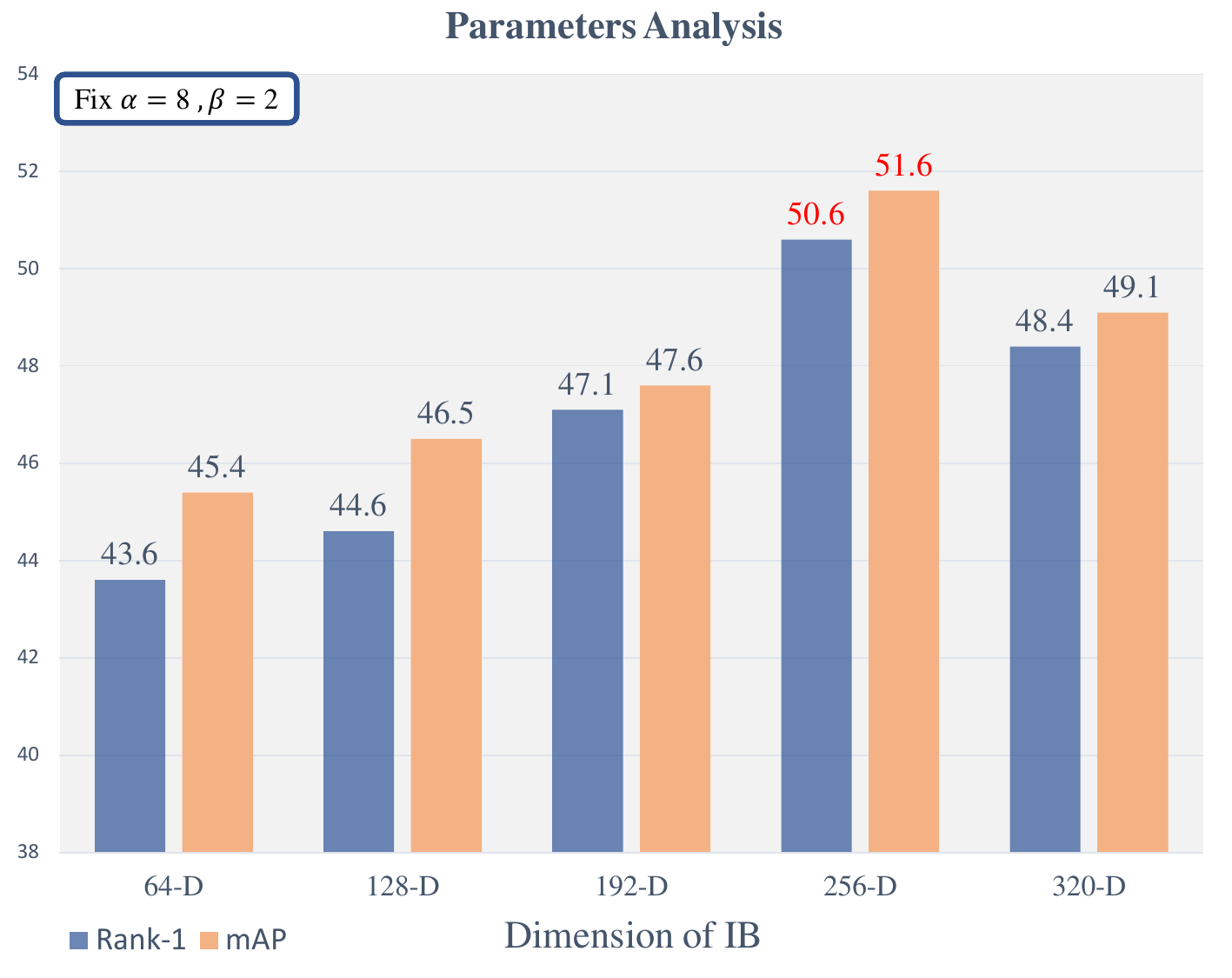}
	}
	\quad
	\subfigure[Accuracy of $v$ when the dimension of IB varies.]{
		\includegraphics[width=0.365\linewidth]{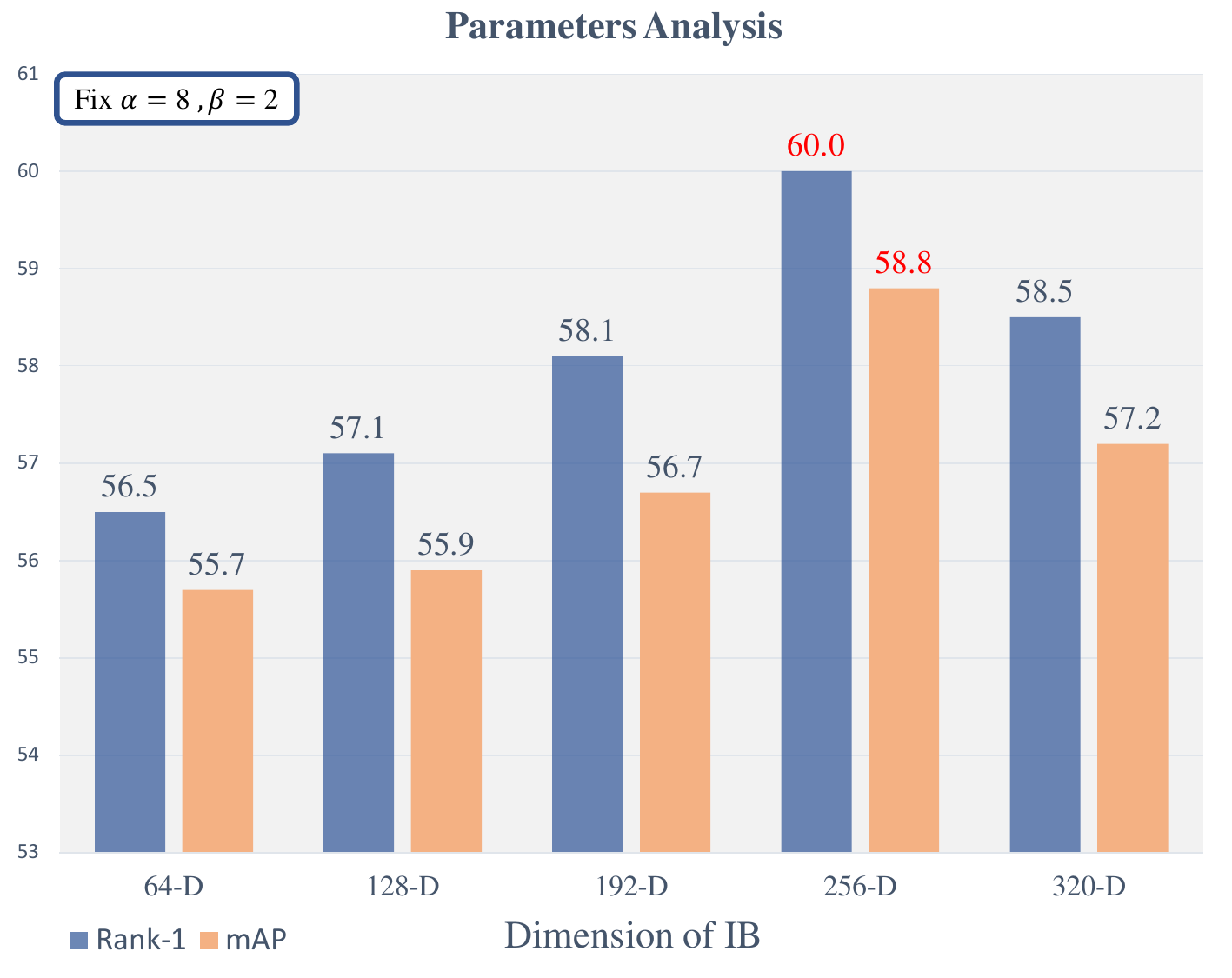}
	}
	\vspace{-2mm}
	\caption{Evaluation of different dimension of the information bottleneck. The evaluation is conducted on SYSU-MM01 under all-search and single-shot mode.}\label{sufficiency fig}
	\vspace{-5mm}
\end{figure}

4) Our approach provides remarkable improvement compared with baseline method (see the 10$^{\text {th}}$-12$^{\text {th}}$ row in Tab. \ref{effectiveness}). By learning predictive cues via the encoding process, it allows both observation and representation to enhance their discriminative ability. However, it is practically impossible to discard all task-irrelevant information without losing any predictive cues, which leads to the performance gap between $v$ and $z$.
In practice, the proposed variational distillation methods could bring about 8.58$\%$ and 5.17$\%$ increments of Rank-1 accuracy for representation $z$ and observation $v$, respectively (compare the 12$^{\text {th}}$ with the 6$^{\text {th}}$ row). Furthermore, comparing with the conventional IB (see the 7$^{\text {th}}$-12$^{\text {th}}$ row), a dramatic accuracy boost of 20.19$\%$@Rank-1 and 17.88$\%$@mAP could be observed.

{\textbf{Sufficiency.}} We also find that the performance gap between $z_{sp}$ and $v_{sp}$ (refer to Fig. \ref{reid framework}) is evidently reduced when using our approach (compare 5$^{\text {th}}$ with 11$^{\text {th}}$ row in Tab. \ref{effectiveness}).
This phenomenon shows that sufficiency of the representation could be better preserved by VSD.
To approximate the minimal sufficient representation, we evaluate VSD by using different dimensions of IB. As is shown in Fig. \ref{sufficiency fig}, we have:

1) When the dimension increases, the accuracy of representation first climbs to a peak, and then degrades. Apparently, if the channel number is extremely reduced, $z$ can not preserve necessary information, while redundant channels are also prone to introduce distractors. Both of them are easy to compromise the accuracy of $z$.

\begin{figure}[t]
	\vspace{-5mm}
	\centering
	\subfigure[$z_{sp}^V$ (VSD)]{
		\includegraphics[width=0.275\linewidth]{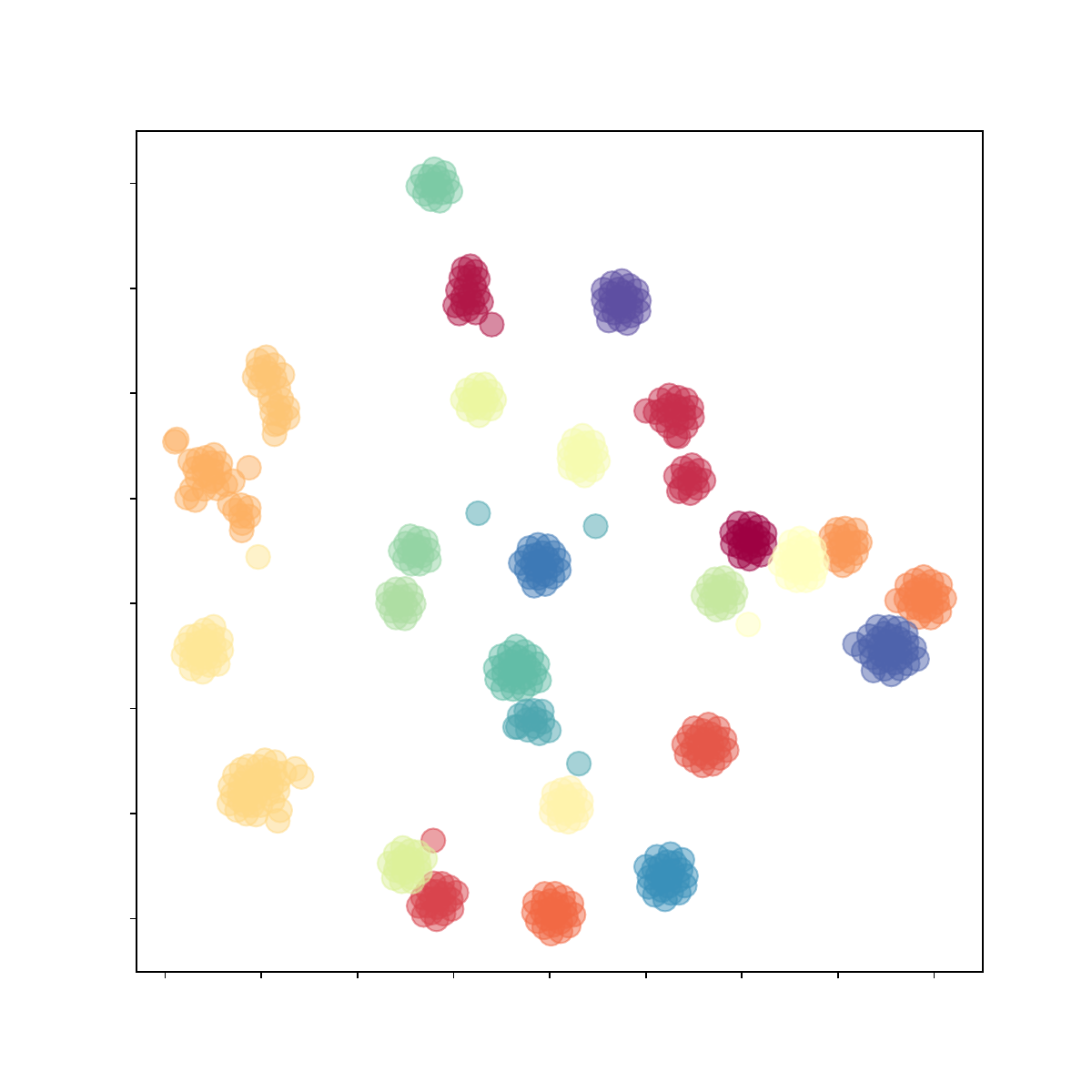}
		\label{6-a}
	}
	\hspace{-7mm}
	\subfigure[$z_{sp}^I$ (VSD)]{
		\includegraphics[width=0.275\linewidth]{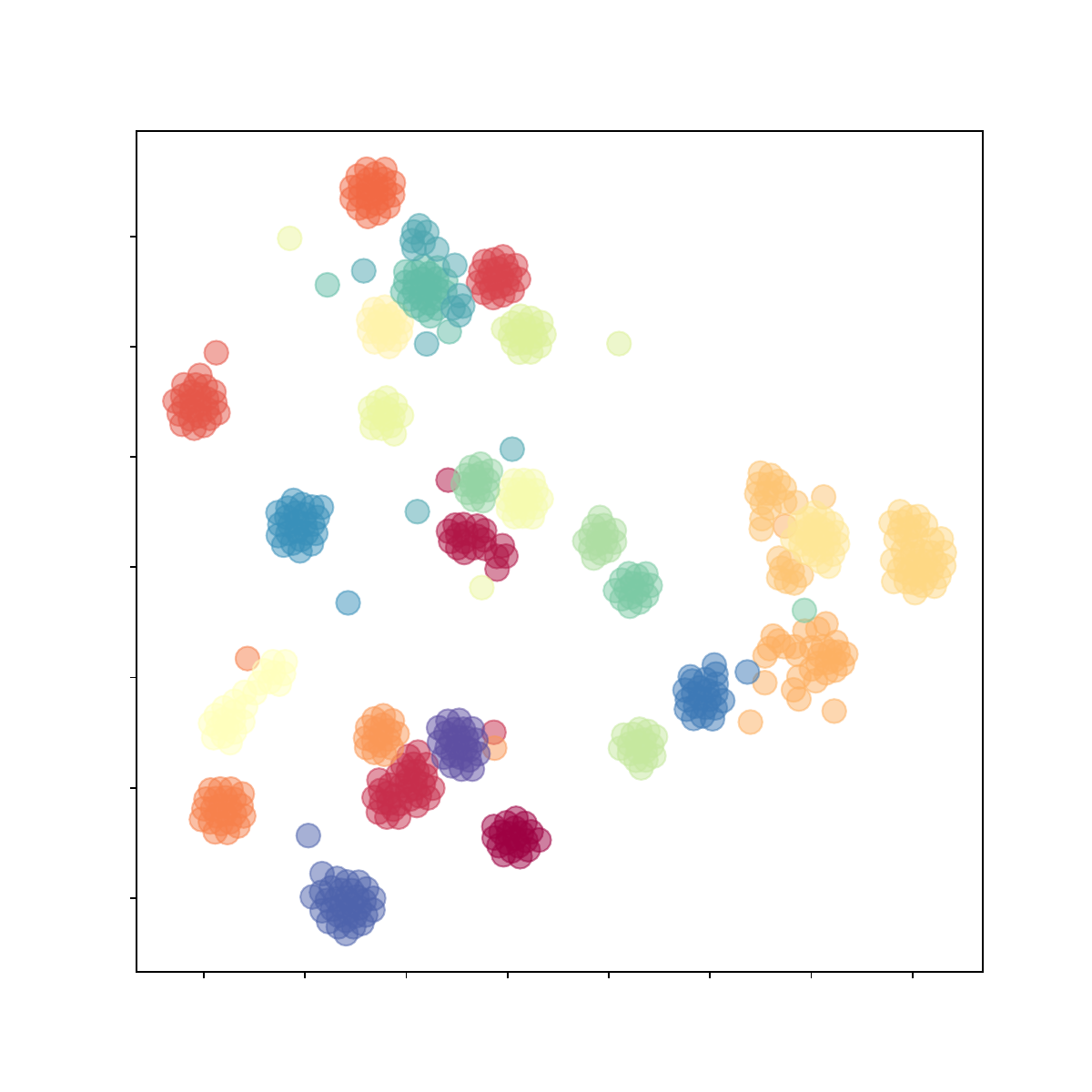}
		\label{6-b}	
	}
	\hspace{-7mm}
	\subfigure[$z_{sh}^V$ (VCD)]{
		\includegraphics[width=0.275\linewidth]{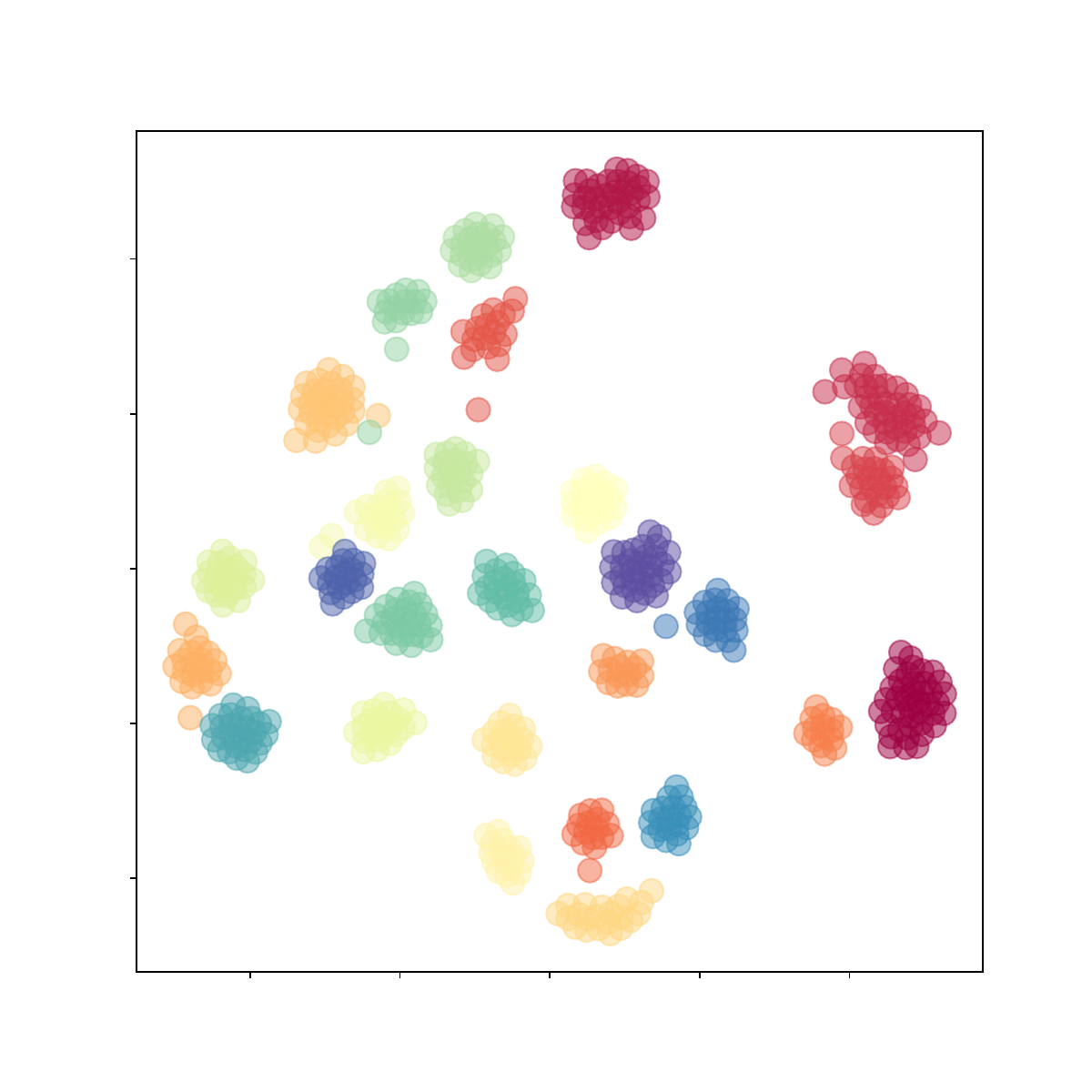}
		\label{6-c}
	}
	\hspace{-7mm}
	\subfigure[$z_{sh}^I$ (VCD)]{
		\includegraphics[width=0.275\linewidth]{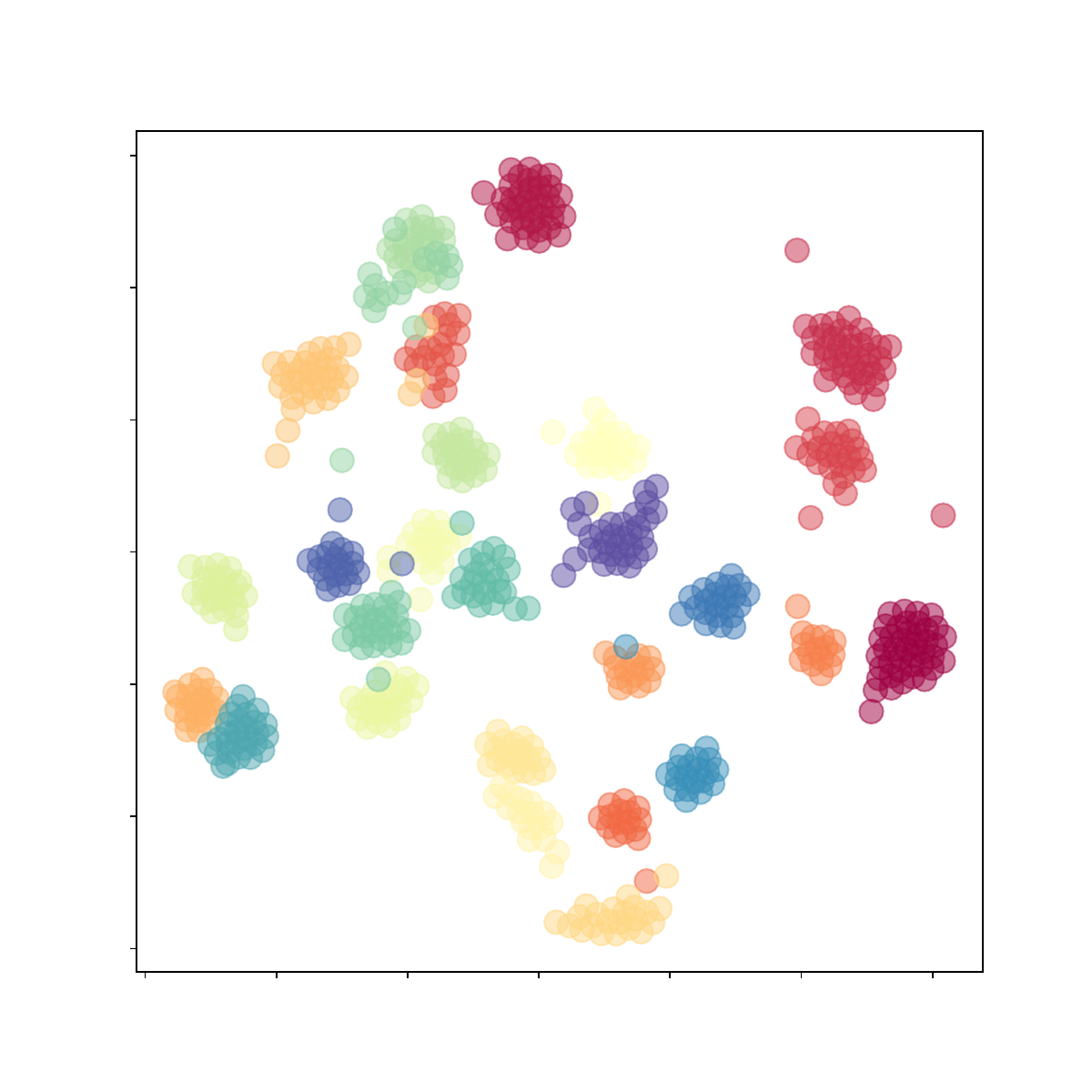}
		\label{6-d}	
	}
	\vspace{-5mm}
	
	\subfigure[$z_{sp}^V$ (CIB)]{
		\includegraphics[width=0.275\linewidth]{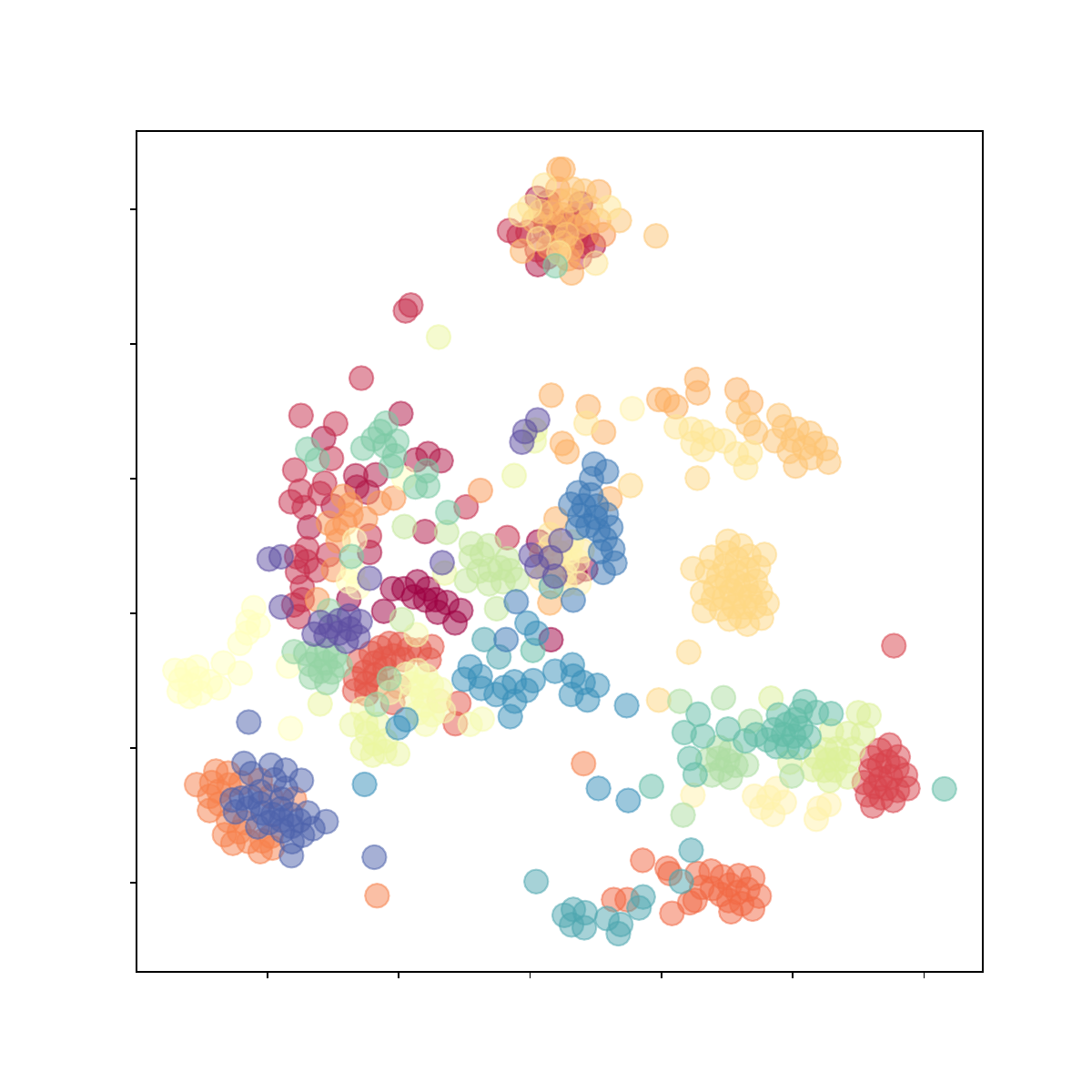}
		\label{6-e}	
	}
	\hspace{-7mm}
	\subfigure[$z_{sp}^I$ (CIB)]{
		\includegraphics[width=0.275\linewidth]{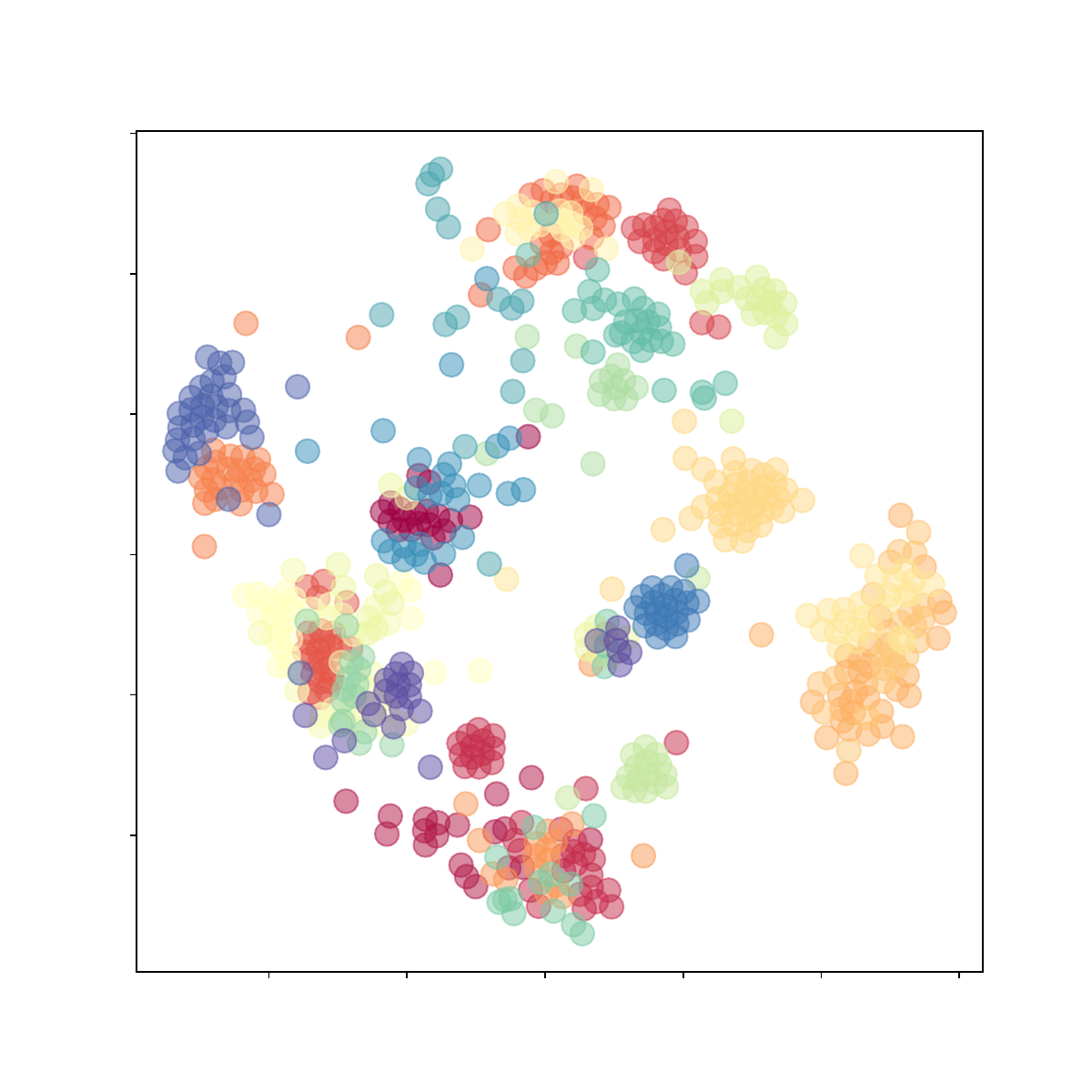}
		\label{6-f}
	}
	\hspace{-7mm}
	\subfigure[$z_{sh}^V$ (CIB)]{
		\includegraphics[width=0.275\linewidth]{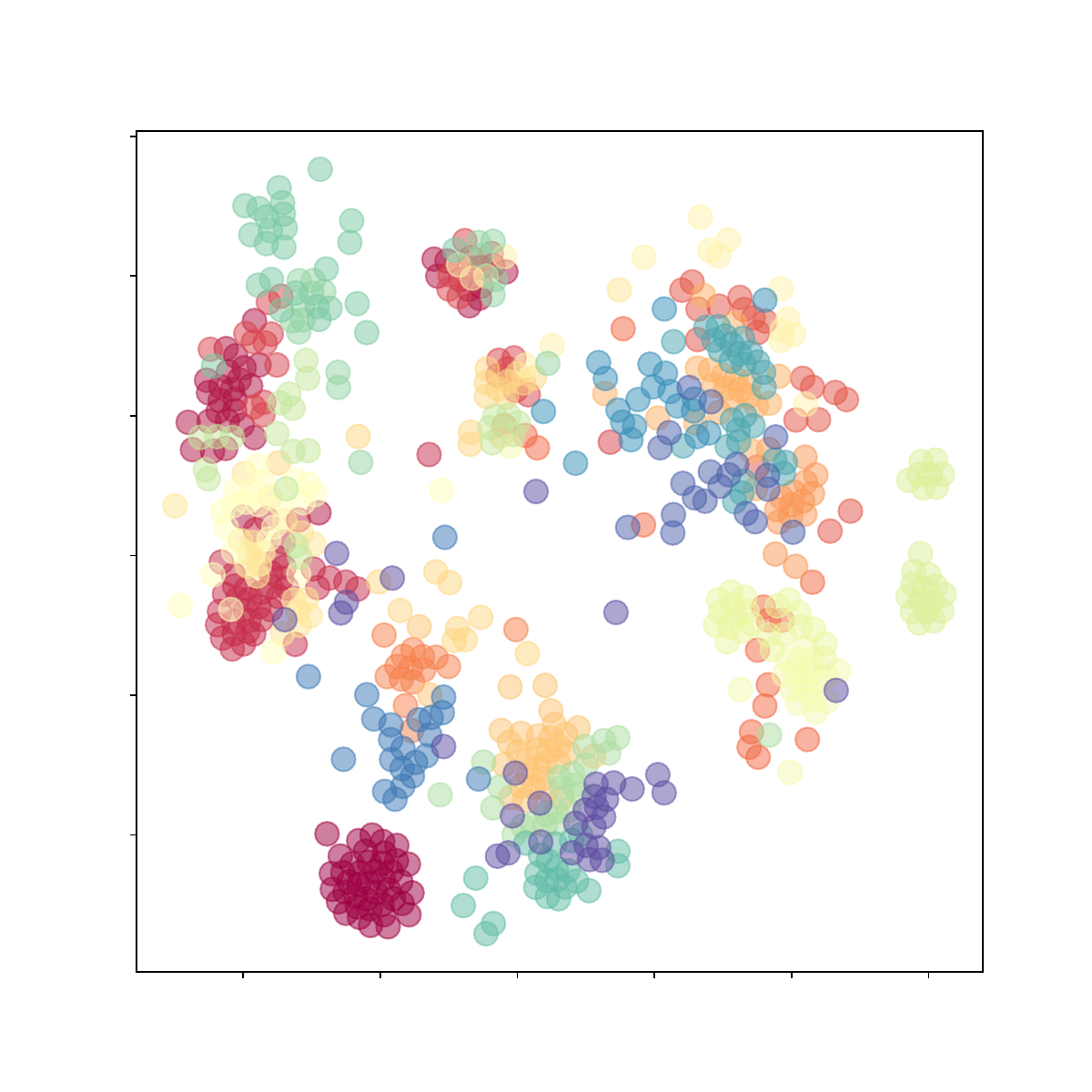}
		\label{6-g}	
	}
	\hspace{-7mm}
	\subfigure[$z_{sh}^I$ (CIB)]{
		\includegraphics[width=0.275\linewidth]{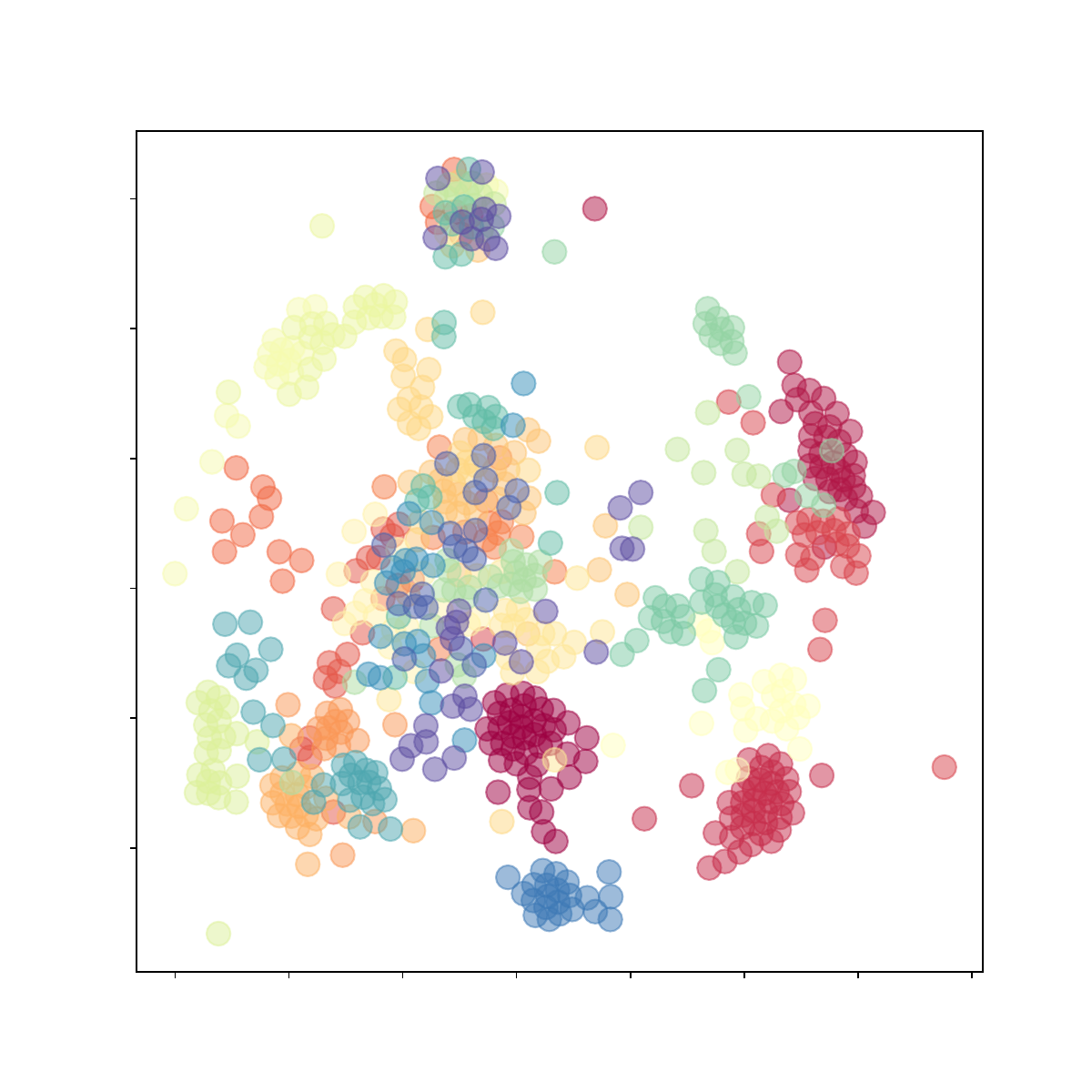}
		\label{6-h}	
	}
	\vspace{-2mm}
	\caption{2D Projection of the embedding space by using t-SNE. The results are obtained from our method and conventional IB on SYSU-MM01 test set. Different colors are used to represent different person IDs.}\label{projection}
	\vspace{-1.6mm}
	\vspace{-0.9mm}
	\vspace{-1.6mm}
	\vspace{-1.6mm}
\end{figure}

2) Different dimensions of $z$ affect the accuracy of the observation. We deduce that the encoder may be implicitly forced to concentrate on the information contained in the representation, and other useful cues are somewhat ignored. 3) The performance gap between $Z$ and $V$ decreases along with the increase of the dimension of $z$. Clearly, the representation is able to hold more useful information with more channels, thus the discrepancy between them is reduced.

We plot the 2D projection of $z_{sp}$ by using t-SNE on Fig. \ref{projection}, where we compare the representations obtained from our approach and conventional IB. As is illustrated in Fig. \ref{6-e} and Fig. \ref{6-f}, the embedding space is mixed in conventional IB, such that the discriminative ability of learned model is naturally limited. On the contrary, our model, which is shown in Fig. \ref{6-a} and Fig. \ref{6-b}, is able to distinguish different pedestrians with the proposed VSD loss, hence we get a substantial improvement whether we use $z$ or $v$.

\begin{figure}[t]
	\centering
	\vspace{-5mm}
	\subfigure[$z_{sh}$ (VCD)]{
		\includegraphics[width=0.365\linewidth]{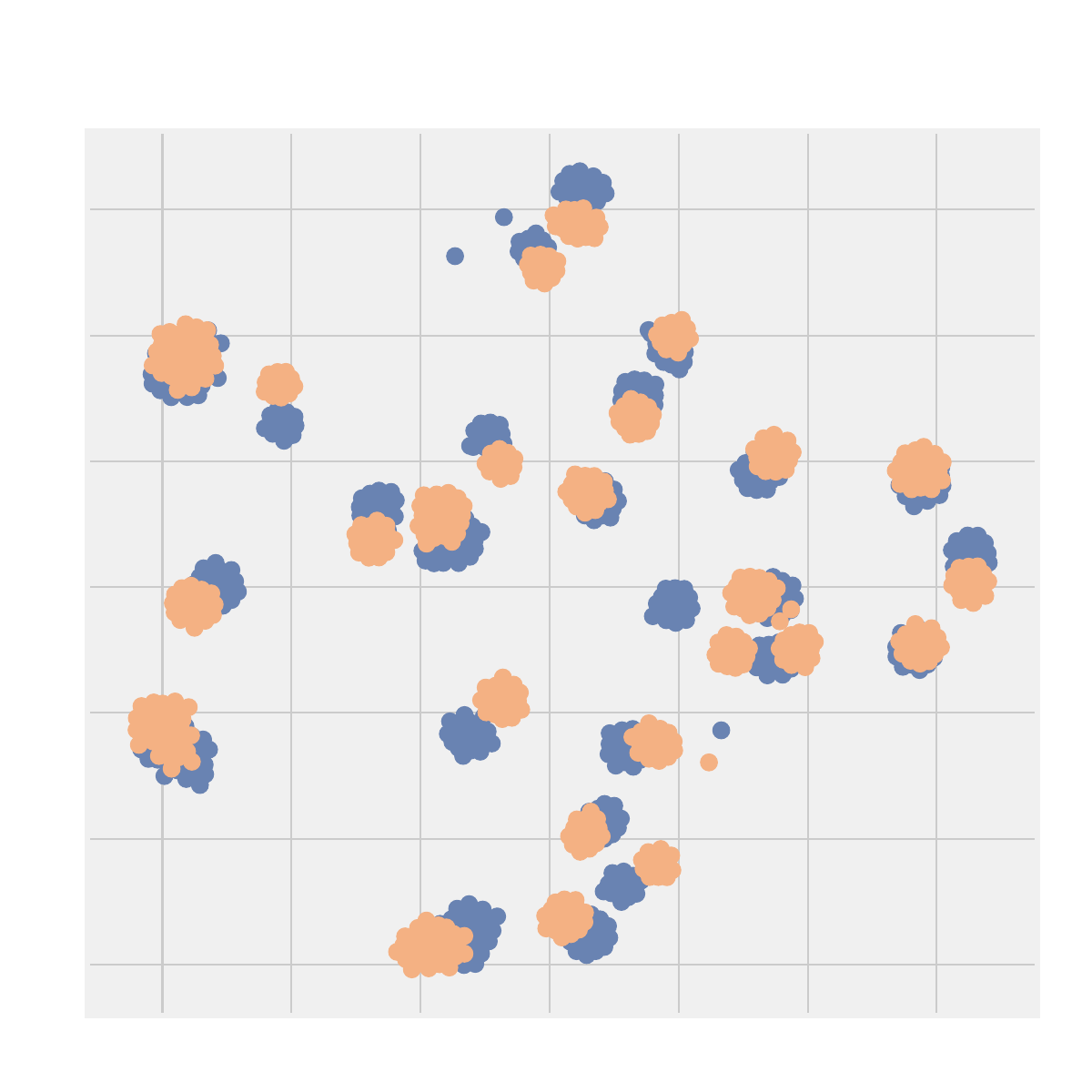}
		\label{fusion_vcd}
	}
	\subfigure[$z_{sh}$ (CIB)]{
		\includegraphics[width=0.365\linewidth]{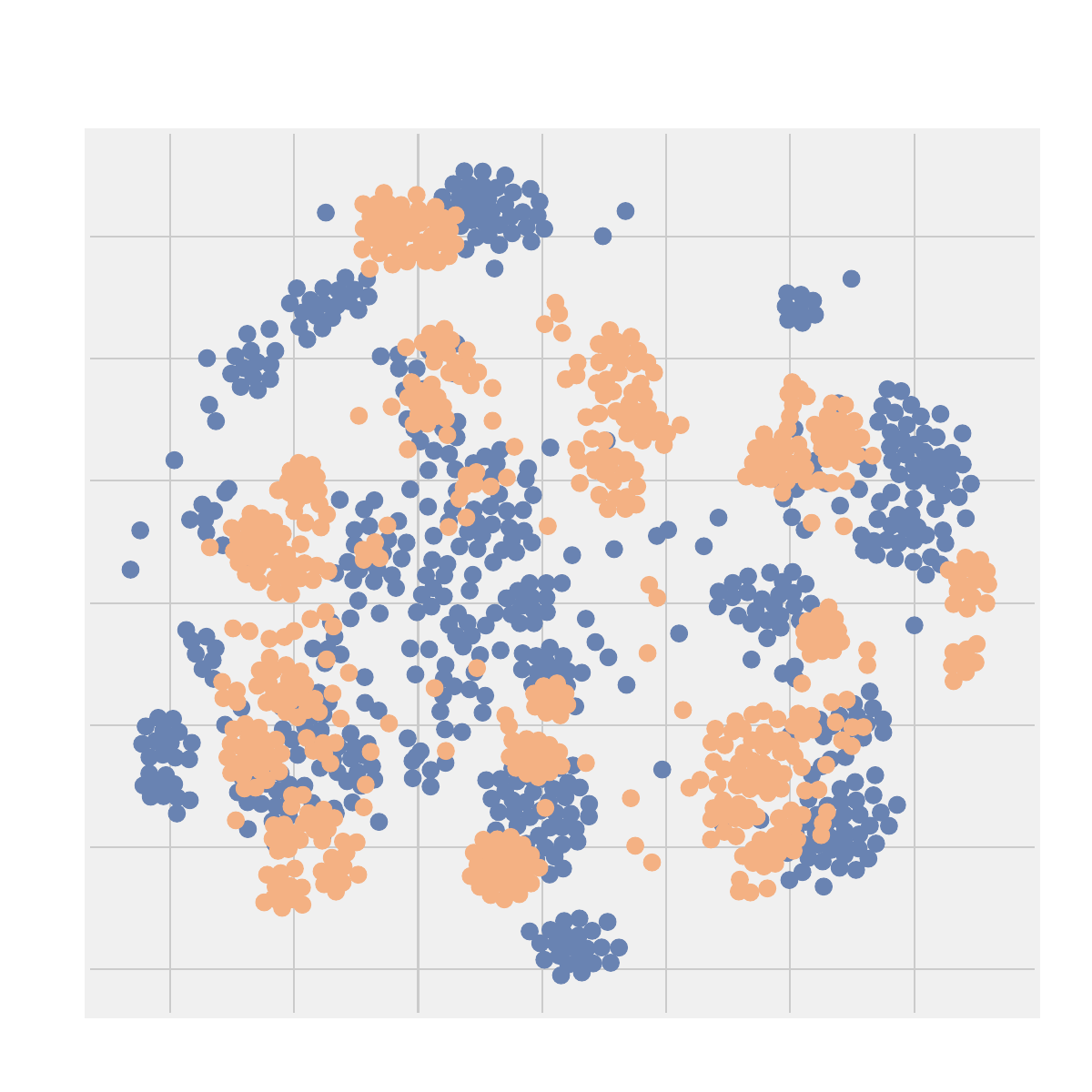}
		\label{fusion_cib}
	}
	\vspace{-2.5mm}
	\caption{2D projection of the joint embedding spaces of $z_{sh}^I$ and $z_{sh}^V$ obtained by using t-SNE on SYSU-MM01.}\label{fusion}
	\vspace{-1.6mm}
	\vspace{-0.9mm}
	\vspace{-1.6mm}
	\vspace{-0.9mm}
	\vspace{-0.5mm}
\end{figure}

{\textbf{Consistency:}} As illustrated in Fig. \ref{projection}, we project the representations obtained from the modal-shared branch onto two principal components using t-SNE. Compared with conventional IB, we make the following observations:

1) From Fig. \ref{6-c} and Fig. \ref{6-d}, we observe that the embedding space corresponding to $z_{sh}^I$ and $z_{sh}^V$ appears to coincide with each other, indicating that most of the view-specific information is eliminated by the proposed method. More importantly, almost all the clusters are distinct and concentrate around a respective centroid, which contrasts with Fig. \ref{6-g} and Fig. \ref{6-h}. This demonstrates the superiority of our approach to the conventional IB.
2) With the comparison between Fig. \ref{6-g} and Fig. \ref{6-h}, we observe that in the embedding space, images of different modals are dramatically discrepant with each other. Such phenomenon is not surprising since conventional IB does not explicitly distinguish view-consistent/specific information. 

To better reveal the elimination of view-specific information, we plot the 2D projection of the joint embedding space of $z_{sh}^I$ and $z_{sh}^V$ in Fig. \ref{fusion}, where the orange and blue circles are used to denote the representation of images of two modals, {\it i.e.}, $x_I$ and $x_V$, respectively. As illustrated in Fig. \ref{fusion_vcd}, clusters of the same identity from different modals are distinct and concentrate around a respective centroid, which contrasts to Fig. \ref{fusion_cib}. This demonstrates the effectiveness of our approach to improve the robustness to view/modal-changes and reduce the discrepancy across modals.

{\textbf{Complexity:}} We also compare the extra computational and memory cost brought by our method and conventional IB. As shown in Table \ref{Complexity}, ``\textbf{Enc}'' denotes the encoder, {\it i.e.}, ResNet-50. ``\textbf{IB}'' denotes the information bottleneck architecture. ``\textbf{MIE}'' denotes the mutual information estimator. Note that the IB is essentially a MLP network with three fully-connected layers. Hence it inevitably introduces extra parameters and computational cost. However, since we use the variation inference to avoid the mutual information estimation, this cost is almost negligible (1.09x training time and 3.04M additional parameters), compared with the heavy mutual information estimator.

\begin{figure}[t]
	\vspace{-2mm}
	\centering
	\subfigure[$\mathcal{L}_{VSD}$]{
		\includegraphics[width=0.4\linewidth]{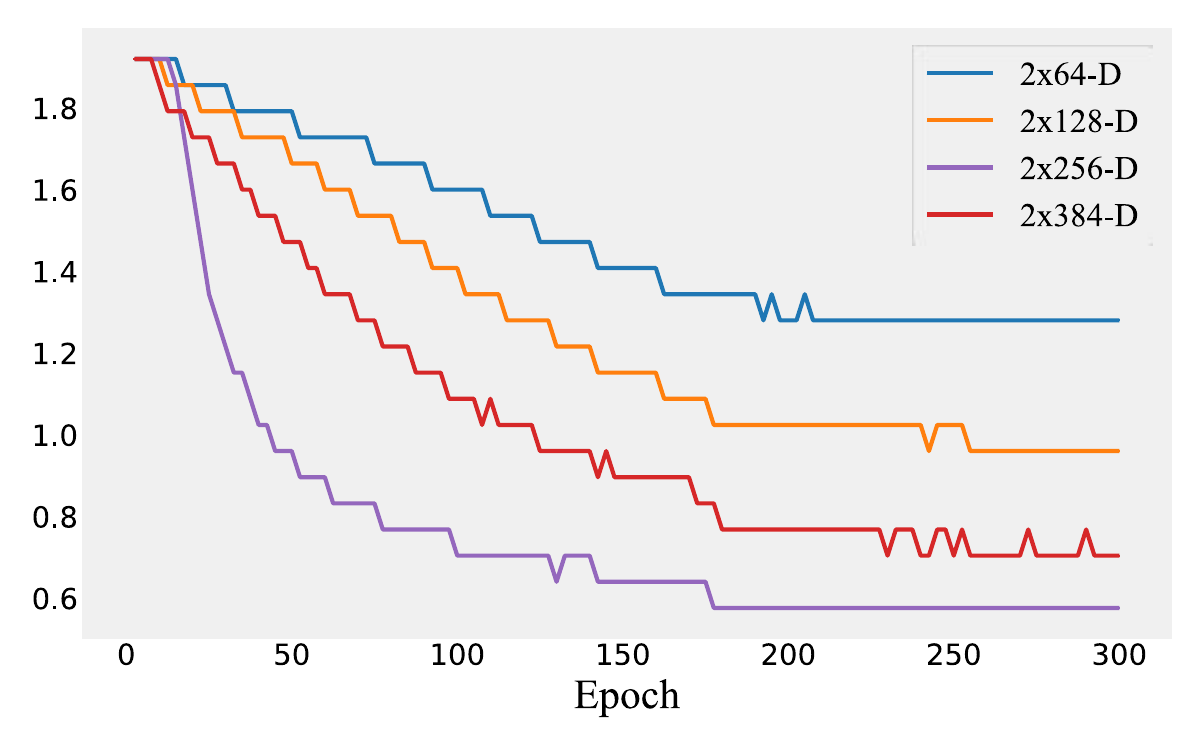}
	}
	\subfigure[$\mathcal{L}_{VCD}$]{
		\includegraphics[width=0.4\linewidth]{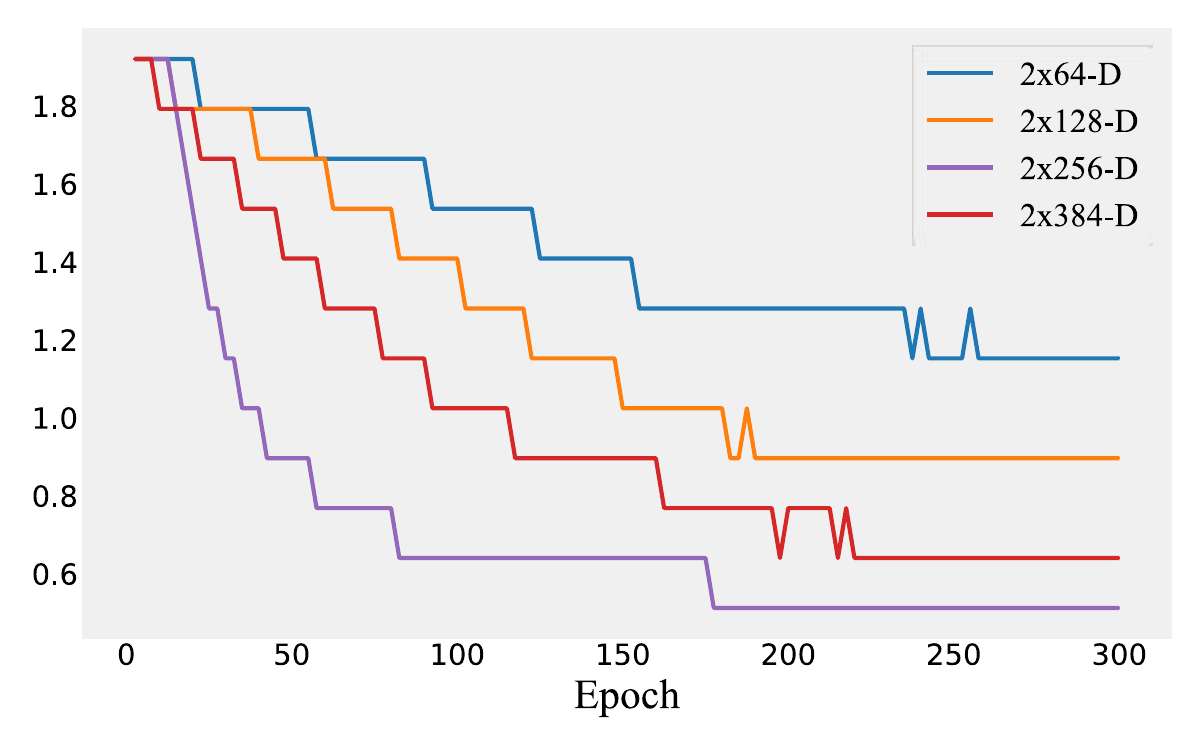}
	}
	\vspace{-1.6mm}
	\vspace{-0.9mm}
	\caption{Evaluation of the discrepancy between $I(v;y)$ and $I(z; y)$, when the dimension of the representation varies. Note the experiments are conduct on SYSU-MM01 dataset under all-search single-shot mode. }\label{curve}
	\vspace{-1.6mm}
	\vspace{-0.9mm}
	\vspace{-1.6mm}
	\vspace{-0.9mm}
\end{figure}

{\textbf{Mutual Information Fitting:}} Fig. \ref{curve} reveals the process of the mutual information fitting between $I(z;y)$ and $I(v;y)$. The amount of predictive information contained in the representation gradually approximates the information contained in observation along the training procedure. Meanwhile, the discrepancy between $I(z;y)$ and $I(v;y)$ varies with the dimension of $z$. Both phenomenons are consistent with our previous observations.

\begin{table}[t]
	\centering
	\renewcommand{\arraystretch}{1.0}
	\small
	\begin{tabular}{|l|lll|l|l|}
		\hline
		Method          & Enc & IB      & MIE     & Time  & Params \\ \hline
		Re-ID Baseline  & $\surd$ &         &         & 1.0x  & 26.37M              \\   
		Ours            & $\surd$ & $\surd$ &         & 1.09x & 29.41M      \\   
		Conventional IB & $\surd$ & $\surd$ & $\surd$ & 1.26x & 35.71M     \\ \hline
	\end{tabular}
	\caption{Computational cost of different methods.}\label{Complexity}
	\vspace{-1.6mm}
	\vspace{-0.9mm}
\end{table}

\section{Conclusion}
In this work, we provide theoretical analysis and obtain an analytical solution to fitting the mutual information by using variational inference, which fundamentally tackle the historical problem. On this basis, we propose Variational Self-Distillation (VSD) and reformulate the objective of IB, which enables us to jointly preserve sufficiency of the representation and get rid of those task-irrelevant distractors. Furthermore, we extend VSD to multi-view representation learning and propose Variational Cross-Distillation (VCD) and Variational Mutual-Learning (VML) to produce view-consistent representations, which are robust to view-changes and result in superior accuracy.
~\\

\textbf{ACKNOWLEDGMENT.}
This work is supported by the National Natural Science Foundation of China 61772524, 61876161, 61972157; National Key Research and Development Program of China (No. 2019YFB2103103); Natural Science Foundation of Shanghai (20ZR1417700); CAAI-Huawei Mind-Spore Open Fund; the Research Program of Zhejiang Lab (No.2019KD0AC02).

{\small
	\bibliographystyle{ieee_fullname}
	\bibliography{egbib}
}
\clearpage

	\section{Appendix}
	\appendix
	In this section, we introduce and prove the theorems mentioned in the main text of this paper. 
	\section{ON SUFFICIENCY}\label{appendix_a}
	
	Consider $x\in\mathbb X$ and $y$ as the input data and the label, and let the $v$ be an observation containing the same amount of predictive information regarding $y$ as $x$ does, and let $z$ be the corresponding representation produced by an information bottleneck. 
	\\\\
	{\textit{Hypothesis}}:
	\\\\
	({\textit{$H_1$}}) $v$ is sufficient for $y$, {\it i.e.}, $I(v;y)=I(x;y)$
	\\\\
	{\textit{Thesis}}:
	\\\\\
	({\textit{$T_1$}}) {\textit{$\min I(v;y)-I(z;y)\iff \min H(y|z)-H(y|v)$}}
	\\\\
	({\textit{$T_2$}}) {\textit{reducing $D_{KL}\left[p(y|v)|| p(y|z)\right]$ is consistent with preserving sufficiency of $z$ for $y$.}}
	\\\\
	{\textit{Proof.}}
	\\
	
	Based on the definition of mutual information \cite{mine}:
	\begin{equation}
		I(v;z):=H(v)-H(v|z),
	\end{equation}
	where $H(v)$ denotes Shannon entropy, and $H(v|z)$ is the conditional entropy of $z$ given $v$ \cite{mine}. Based on the symmetry of mutual information, we have:
	\begin{equation}
		I(v;z)=I(z;v),
	\end{equation}
	which indicates that the requirement of sufficiency is equivalent to:
	\begin{flalign}
		&I(v;y)=I(z;y)&\nonumber\\
		&\iff I(y;v)=I(y;z)&\nonumber\\
		&\iff H(y)-H(y|v)=H(y)-H(y|z)&\nonumber\\
		&\iff -H(y|v)=-H(y|z).&\label{deduction1}
	\end{flalign}
	Therefore, we have: 
	\begin{equation}
		\min I(v;y)-I(z;y)\iff \min H(y|z)-H(y|v),
	\end{equation}
	which proves ({\textit{$T_1$}}). Based on the definition of conditional entropy, for any continuous variables $v,y$ and $z$, we have:
	\begin{flalign} 
		&I(v;y)-I(z;y)=H(y|z)-H(y|v)=& \nonumber\\
		&-\int p(z)dz\int p(y|z)\log p(y|z)dy\nonumber\\
		&+\int p(v)dv\int p(y|v)\log p(y|v)dy~=\nonumber\\
		&-\iint p(z) p(y|z)  \log \left[\frac{p(y|z)}{p(y|v)}  p(y|v)\right] dzdy\nonumber\\
		&+\iint p(v) p(y|v)  \log \left[\frac{p(y|v)}{p(y|z)}  p(y|z)\right] dvdy.\label{proof1}
	\end{flalign}
	
	By factorizing the double integrals in Eq. (\ref{proof1}) into another two components, we show the following:
	\begin{flalign} 
		&\iint p(z) p(y|z)  \log \left[\frac{p(y|z)}{p(y|v)}  p(y|v)\right] dzdy=&\nonumber\\\nonumber\\
		&\iint \underbrace{p(z) p(y|z)  \log \frac{p(y|z)}{p(y|v)} dzdy}_{\operatorname{term} Z_{1}}~+\nonumber\\
		&\iint\underbrace{p(z)p(y|z)\log p(y|v) dzdy}_{\operatorname{term} Z_{2}}\label{proof2}.
	\end{flalign}
	
	Conduct similar factorization for the second term in Eq.(\ref{proof1}), we have:
	\begin{flalign} 
		&\iint p(v) p(y|v)  \log \left[\frac{p(y|v)}{p(y|z)}  p(y|z)\right] dvdy=&\nonumber\\\nonumber\\
		&\iint \underbrace{p(v) p(y|v)  \log \frac{p(y|v)}{p(y|z)} dvdy}_{\operatorname{term} V_{1}}~+\nonumber\\
		&\iint\underbrace{p(v)p(y|v)\log p(y|z) dvdy}_{\operatorname{term} V_{2}}\label{proof3}.
	\end{flalign}

	Integrate term $Z_1$ and term $V_1$ over $y$:
	\begin{equation}
		Z_{1}=\int p(z)  D_{KL}[p(y|z) \| p(y|v)] dz,
	\end{equation}
	\begin{equation}
		V_{1}=\int p(v)  D_{KL}[p(y|v) \| p(y|z)] dv,
	\end{equation}
	where $D_{KL}$ denotes KL-divergence. Integrate term $Z_2$ and term $V_2$ over $z$ and $v$ respectively, we have:
	\begin{equation}
		Z_2=\int p(y)  \log p(y|v) dy.
	\end{equation}
	\begin{equation}
		V_2=\int p(y)  \log p(y|z) dy
	\end{equation}
	
	In the view of above, we have the following:
	\begin{flalign}
		&I(v;y)-I(z;y) = H(y|z)-H(y|v)=&\nonumber\\
		&\int p(v)  D_{KL}[p(y|v) \| p(y|z)]dv + \int p(y)  \log \left[\frac{p(y|z)}{p(y|v)}\right]dy \nonumber\\
		&- \int p(z)  D_{KL}[p(y|z) \| p(y|v)] dz \label{sufficiency object}
	\end{flalign}
	
	Based on the non-negativity of KL-divergence, Eq. (\ref{sufficiency object}) is upper bounded by:
	\begin{flalign}
		&\int p(v)  D_{KL}[p(y|v) \| p(y|z)]dv + \int p(y)  \log \left[\frac{p(y|z)}{p(y|v)}\right]dy.
	\end{flalign}
	
	Equivalently, we have the upper bound as: 
	\begin{flalign}
		&\mathbb{E}_{v\sim E_{\theta}(v|x)}\mathbb{E}_{z\sim E_{\phi}(z|v)}[D_{KL}[p(y|v)\|p(y|z)]] &\nonumber\\
		&+\mathbb{E}_{v\sim E_{\theta}(v|x)}\mathbb{E}_{z\sim E_{\phi}(z|v)}\left[\log \left[\frac{p(y|z)}{p(y|v)}\right]\right],
	\end{flalign}
	where $\theta,\phi$ denote the parameters of the encoder and the information bottleneck.
	Therefore, the objective of preserving sufficiency of $z$ for y can be formalized as:
	\begin{flalign}
		&\min_{\theta, \phi} \mathbb{E}_{v\sim E_{\theta}(v|x)}\mathbb{E}_{z\sim E_{\phi}(z|v)}\left[D_{K L}[\mathbb{P}_v||\mathbb{P}_z]+ \log \left[\frac{\mathbb{P}_z}{\mathbb{P}_v}\right]\right],\label{sufficiency_loss}&
	\end{flalign}
	in which $\mathbb{P}_z=p(y|z)$ and $\mathbb{P}_v=p(y|v)$ denote the predicted distributions of the representation and observation.
	
	Clearly, the objective of preserving sufficiency is equivalent to minimize the discrepancy between the predicted distributions of $v$ and $z$. Notice that this can be achieved by minimizing $D_{KL}(\mathbb{P}_v||\mathbb{P}_z)$, which can explicitly approximate $p(y|z)$ to $p(y|v)$ and implicitly reduce the second term in Eq.(\ref{sufficiency_loss}) in the same time. At the extreme, the representation $z$ retrieves all label information contained in the sufficient observation $v$, indicating that $z$ is sufficient for $y$ as well. Formally, we have:
	\begin{equation}
		\lim _{\mathbb{P}_z \rightarrow\mathbb{P}_v} D_{KL}[\mathbb{P}_v|| \mathbb{P}_z] + \int p(y)  \log \left[\frac{\mathbb{P}_v}{\mathbb{P}_z}\right] dy = 0
	\end{equation}
	Based on Eq. (\ref{sufficiency object}) , we show the following:
	\begin{equation}
		\lim _{\mathbb{P}_z \rightarrow\mathbb{P}_v} I(v;y)-I(z;y)=\lim _{\mathbb{P}_z \rightarrow\mathbb{P}_v}H(y|v)-H(y|z)=0
	\end{equation}
	which reveals that minimizing $D_{KL}[\mathbb{P}_v||\mathbb{P}_z]$ is consistent with the objective of preserving sufficiency of the representation. Thus ({\textit{$T_2$}}) holds.


	\section{ON CONSISTENCY}\label{appendix_b}
	Consider $v_1,v_2$ as two sufficient observations of the same objective $x$ from different viewpoints or modals, and let $y$ be the label. Let $z_1,z_2$ be the corresponding representations obtained from an information bottleneck. 
	\\\\
	{\textit{Hypothesis}}:
	\\\\
	({\textit{$H_1$}}) both $v_1,v_2$ are sufficient for $y$
	\\\\
	({\textit{$H_2$}}) $z_1,z_2$ are in the same distribution
	\\\\
	{\textit{Thesis}}:
	\\\\
	({\textit{$T_1$}}) minimizing $D_{KL}[p(y|v_2)\|p(y|z_1)]$ is consistent with the objective of eliminating task-irrelevant information encoded in $I(z_1;v_2)$, and is able to preserve those predictive and view-consistent information, vice versa for $D_{KL}[p(y|v_1)\|p(y|z_2)]$ and $I(z_2;v_1)$
	\\\\
	({\textit{$T_2$}}) minimizing $D_{JS}[p(y|z_1)\|p(y|z_2)]$ is consistent with the objective of elimination of view-specific information for both $z_1$ and $z_2$
	\\\\
	({\textit{$T_3$}}) performing VCD and VML can promote view-consistency between $z_1$ and $z_2$
	\\\\
	{\textit{Proofs.}}
	\\
	
	By factorizing the mutual information between the data observation $v_1$ and its representation $z_1$, we have:
	\begin{equation}
		I(v_1;z_1)=I(v_1;z_1|v_2)+I(z_1;v_2),\label{factorization2}
	\end{equation}
	where $I(z_1;v_2)$ and $I(v_1;z_1|v_2)$ denote the view-consistent and view-specific information, respectively. 
	
	Furthermore, by using the chain rule of mutual information, which subdivides $I(z_1;v_2)$ into two components (proofs could be found in \cite{mib}), we have:
	\begin{equation}
		I(z_1;v_2)=I(v_2;z_1|y)+I(z_1;y)
	\end{equation}
	combining with Eq. (\ref{factorization2}), we show the following:
	\begin{equation}
		I(z_1;v_1)=\underbrace{I(v_1;z_1|v_2)}_{\operatorname{view-specific}}+\underbrace{I(v_2;z_1|y)}_{\operatorname{superfluous}}+\underbrace{I(z_1;y)}_{\operatorname{predictive}},
	\end{equation}
	
	Based on Appendix \ref{appendix_a}, reducing $D_{KL}[\mathbb{P}_{v_2}||\mathbb{P}_{z_1}]$, where $\mathbb{P}_{z_1}=p(y|z_1),\mathbb{P}_{v_2}=p(y|v_2)$, can minimize $I(v_2;z_1|y)$ and maximize $I(z_1;y)$ in the same time, thus we conclude that ({\textit{$T_1$}}) holds.

	Considering that $z_1,z_2\in\mathbb{Z}$, $I(v_1;z_1|v_2)$ can be expressed as:
	\begin{flalign}
		&I(v_1;z_1|v_2)=\mathbb{E}_{v_1,v_2\sim E_{\theta}(v|x)}\mathbb{E}_{z_1,z_2\sim E_{\phi}(z|v)}\left[\log \frac{p(z_1|v_1)}{p(z_1|v_2)}\right]&\nonumber\\
		&=\mathbb{E}_{v_1,v_2\sim E_{\theta}(v|x)}\mathbb{E}_{z_1,z_2\sim E_{\phi}(z|v)}\left[\log \frac{p(z_1|v_1) p(z_2|v_2)}{p(z_2|v_2) p(z_1|v_2)}\right]\nonumber\\
		&=D_{KL}[p(z_1|v_1)||p(z_2|v_2)]-D_{KL}[p(z_2|v_1)||p(z_2|v_2)]\nonumber\\
		&\le D_{KL}[p(z_1|v_1)||p(z_2|v_2)].\label{vml deduction}
	\end{flalign}
	
	Notice this bound is tight whenever $z_1$ and $z_2$ produce consistent encodings \cite{mib}, which can be assured by the proposed VCD and is visualized in the main body of this paper.
	On the other hand, since $y$ is constant with respect to the parameters to be optimized, we utilize Eq. (\ref{vml_1}) to approximate Eq. (\ref{vml deduction}):
	\begin{equation}
		\mathbb{E}_{v_1,v_2\sim E_{\theta}(v|x)}\mathbb{E}_{z_1,z_2\sim E_{\phi}(z|v)}\left[D_{KL}[\mathbb{P}_{z_1}||\mathbb{P}_{z_2}]\right],\label{vml_1}
	\end{equation}
	in which $\mathbb{P}_{z_1}=p(y|z_1)$ and $\mathbb{P}_{z_2}=p(y|z_2)$ denote the predicted distributions. Based on the above analysis, we conclude that $I(v_1;z_1|v_2)$ can be minimized by reducing $D_{KL}[\mathbb{P}_{z_1}||\mathbb{P}_{z_2}]$. Similarly, we introduce the following objective to minimize $I(v_2;z_2|v_1)$.
	\begin{equation}
		\mathbb{E}_{v_1,v_2\sim E_{\theta}(v|x)}\mathbb{E}_{z_1,z_2\sim E_{\phi}(z|v)}\left[D_{KL}[\mathbb{P}_{z_2}||\mathbb{P}_{z_1}]\right],\label{vml_2}
	\end{equation}
	For simplicity, we apply Eq. (\ref{vml_3}) to eliminate the view-specific information for both $z_1$ and $z_2$.
	\begin{equation}
		\min_{\theta, \phi} \mathbb{E}_{v_1,v_2\sim E_{\theta}(v|x)}\mathbb{E}_{z_1,z_2\sim E_{\phi}(z|v)}\left[D_{JS}[\mathbb{P}_{z_1}||\mathbb{P}_{z_2}]\right],\label{vml_3}
	\end{equation}
	where $D_{JS}$ denotes the Jensen-Shannon divergence. Thus ({\textit{$T_2$}}) holds.
	
	Finally, according to \cite{mib}, $I(z_1;y)=I(v_1v_2;y)$ when the following hypotheses stand: $z_1$ is a representation of $v_1$ and $I(y;z_1|v_1v_2)=0$, both $v_1$ and $v_2$ are sufficient for $y$, $z_1$ is sufficient for $v_2$. As a consequence of data processing inequality, the amount of information encoded in $z_1$ cannot be more than the joint observation, {\it i.e.} $I(y;z_1|v_1v_2)\equiv 0$. Since sufficiency of $v_1$ and $v_2$ for $y$ is consistent with the given task, it is widely adopted as an established assumption. Notably, sufficiency of $z_1$ for $v_2$ can be achieved by preserving view-consistent information while simultaneously eliminating the view-specific details, which correspond to the proposed VCD and VML, respectively. Therefore, ({\textit{$T_3$}}) holds.

\end{document}